\documentclass[]{fairmeta}
\usepackage{microtype}
\usepackage{lmodern}
\usepackage{amsfonts}
\usepackage{graphicx}
\usepackage{booktabs}
\usepackage{wrapfig}
\usepackage{amsmath}
\usepackage{amsthm}
\usepackage{algpseudocode}
\usepackage[linesnumbered,lined,boxed,commentsnumbered,ruled,longend]{algorithm2e}
\usepackage[capitalize,noabbrev]{cleveref}
\theoremstyle{plain}

\theoremstyle{definition}

\theoremstyle{remark}

\usepackage{enumitem}
\title{\raisebox{-3pt}{\includegraphics[width=0.045\linewidth]{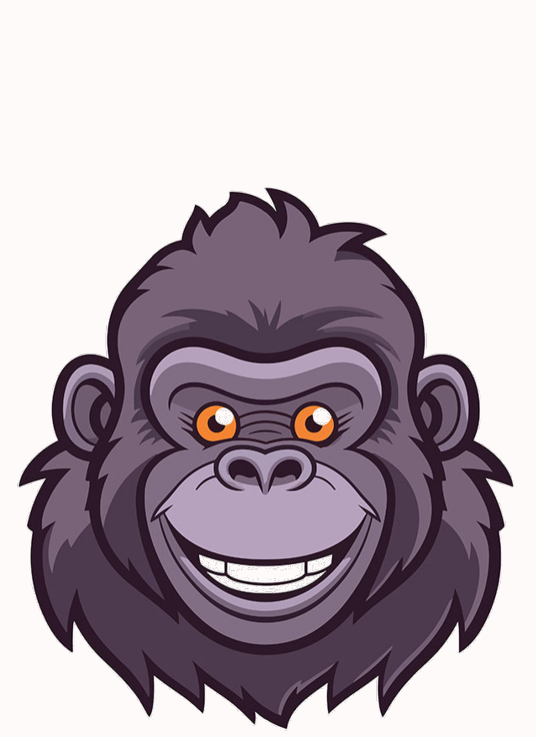}}
APE: Faster and Longer Context-Augmented Generation via \underline{A}daptive \underline{P}arallel \underline{E}ncoding}

\author[\dagger]{Xinyu Yang}
\author[\dagger \ddagger] {Tianqi Chen}
\author[\dagger]{Beidi Chen}

\affiliation[\dagger]{Carnegie Mellon University}
\affiliation[\ddagger]{Nvidia}

\abstract{
 Context-augmented generation (CAG) techniques, including RAG and ICL, require the efficient combination of multiple contexts to generate responses to user queries. Directly inputting these contexts as a sequence introduces a considerable computational burden by re-encoding the combined selection of contexts for every request. To address this, we explore the promising potential of parallel encoding to independently pre-compute and cache each context's KV states. This approach enables the direct loading of cached states during inference while accommodating more contexts through position reuse across contexts. However, due to misalignments in attention distribution, directly
applying parallel encoding results in a significant performance drop. To enable effective and efficient CAG, we propose \underline{A}daptive \underline{P}arallel \underline{E}ncoding (\textbf{APE}), which brings shared prefix, attention temperature, and scaling factor to align the distribution of parallel encoding with sequential encoding. Results on RAG and ICL tasks demonstrate that APE can preserve 98\% and 93\% sequential encoding performance using the same inputs while outperforming parallel encoding by 3.6\% and 7.9\%, respectively. It also scales to many-shot CAG, effectively encoding hundreds of contexts in parallel. Efficiency evaluation shows that APE can achieve an end-to-end 4.5$\times$ speedup by reducing 28$\times$ prefilling time for a 128K-length context.
}

\metadata[Github]{\url{https://github.com/Infini-AI-Lab/APE}}
\metadata[Website]{\url{https://infini-ai-lab.github.io/APE-Page}}

\begin{document}

\maketitle
\section{Introduction}
\begin{figure}[t]
    \centering
    \includegraphics[width=\textwidth]{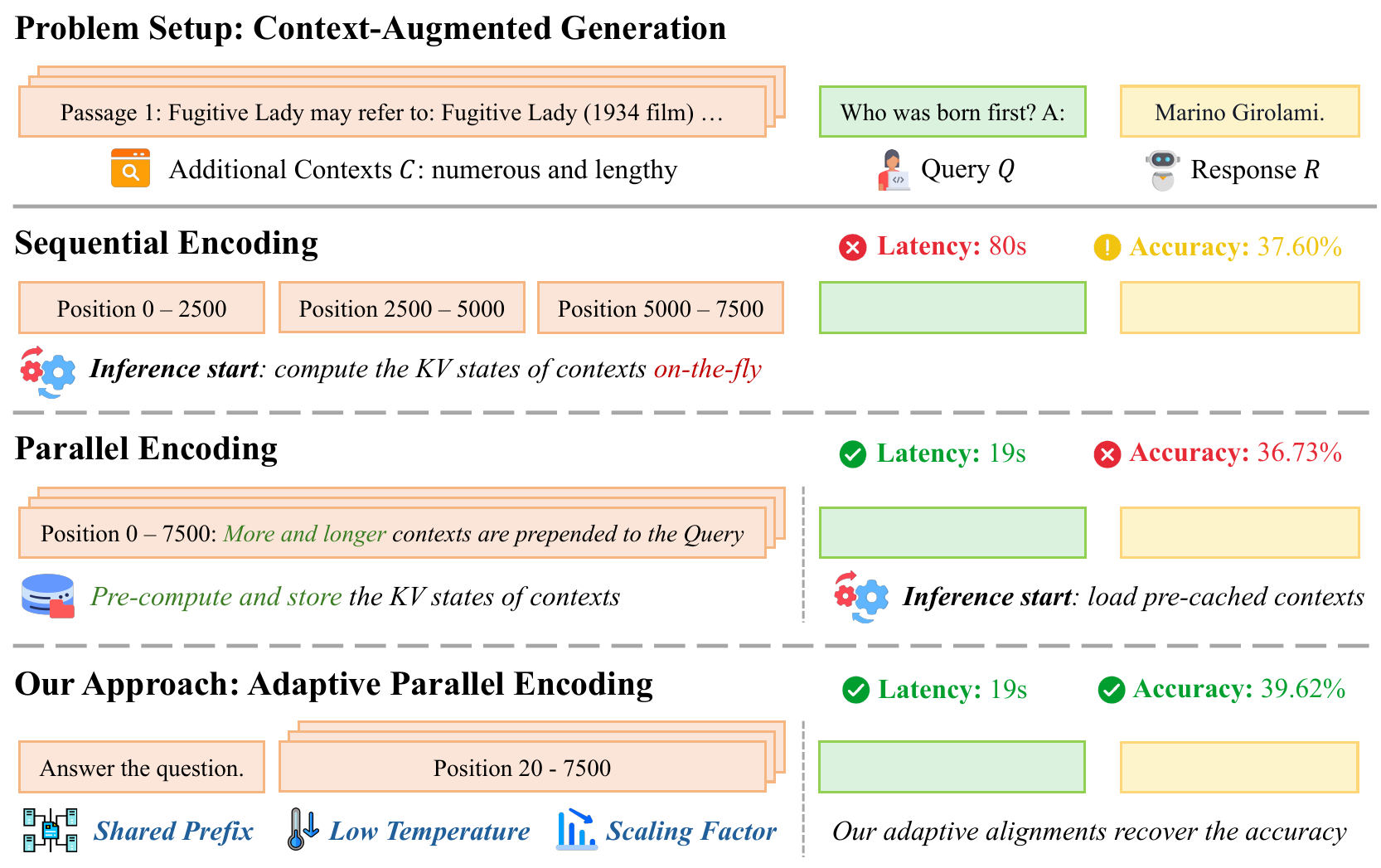}
    \caption{
    \textbf{Overview of Our Approach.} Context-augmented generation leverages additional contexts to improve LLM response quality to user queries. 
    Sequential encoding prefills selected context chunks as a long sequence during inference, leading to high latency from on-the-fly re-encoding and low accuracy due to context window limitations.
    Parallel encoding offers an alternative method to pre-compute more and longer contexts within the same positional range but results in worse performance.
    To address these challenges, we propose \underline{A}daptive \underline{P}arallel \underline{E}ncoding (\textbf{APE}) to re-align the attention weight distribution of parallel encoding with sequential encoding via three training-free steps: shared prefix, scaling factor, and adaptive temperature, leading to fast and accurate CAG systems in real-world applications.}
    \label{fig:intro}
\end{figure}

Recent advances in context-augmented generation (CAG) techniques, particularly retrieval-augmented generation (RAG)~\citep{gupta2024rag, gao2023retrieval} and in-context learning (ICL)~\citep{dong2022survey, wei2022emergent}, have been widely adopted in large language models (LLMs)~\citep{llama3, achiam2023gpt}, improving their ability to generalize to unseen tasks with contextual information, as demonstrated in Figure~\ref{fig:intro} (top).
These techniques employ a \textit{sequential encoding} process to ground LLM inputs with knowledge from external sources: concatenating the retrieved texts into one sequence, and encoding the sequence into key-value (KV) states as the context for subsequent queries. While this new, significantly longer input improves performance, the increased latency in context prefilling becomes a bottleneck in tasks that require long inputs but generate short outputs~\citep{bai2023longbench, agarwal2024many, jiang2024longrag}. For example, prefilling a 128K context takes 17 seconds, whereas generating 256 tokens requires only 6 seconds. This discrepancy leaves significant room to improve the practical efficiency of CAG systems in real-world deployments~\citep{Liu_LlamaIndex_2022, Chase_Longchain_2022}.

Since texts for CAG are typically stored independently in external databases~\citep{Qdrant, douze2024faiss}, pre-caching all these texts for direct loading during inference offers a brute-force approach to accelerate CAG. However, for autoregressive LLMs, the KV states are inherently context-dependent. This dependency makes naive pre-caching impractical, as it would require caching all possible context permutations, leading to factorial growth in memory requirements as the database size increases. For instance, caching all permutations of just ten 256-token text chunks for the \textsc{LLaMA-3-8B} model would demand an impractical 22 PB of memory.

To address this issue, \textit{parallel encoding}~\citep{ratner2022parallel, yen2024long, li2024focusllm, Sun2024BlockAttentionFE} is introduced to encode each context into KV states separately, ensuring that tokens from different contexts cannot attend to each other during encoding. Next, the on-the-fly generation starts by prefilling user queries, which can attend to the cached KV states from all contexts without re-encoding, offering two benefits:

\textbf{Pre-caching Contexts for Fast Inference:} Texts from external sources can be pre-computed and cached into KV states, which serve as contexts for direct loading during inference. Additionally, this approach allows for cost-free manipulation of contexts, including operations like insertion, deletion, replacement, and swapping.

\textbf{Re-using Positions for Long Context:} Contexts can be inserted into the same range of positions in an LLM's context window, allowing for more and longer context chunks. It also mitigates the problem of ``lost in the middle" in context ordering~\citep{liu2024lost}, as each context is equally ``close'' to the generated tokens. 

Despite these advantages, parallel encoding leads to significant performance degradation across multiple RAG and ICL scenarios, as shown in Figure \ref{fig:obseravtion1}, with average declines of 4.9\% (despite using 2-10$\times$ more contexts) and 49.0\%, respectively. While prior works~\citep{Sun2024BlockAttentionFE, yen2024long} have attempted to correct this with fine-tuning, these methods continue to exhibit reduced accuracy in reasoning tasks (e.g., GSM8K). This decrease arises from the limited generalization capability of models fine-tuned on simple tasks to complex ones.

However, our results in Figure~\ref{fig:obseravtion1} also reveal that parallel encoding holds promise, as LLMs can still generate reasonable responses due to their inherent alignments with sequential encoding. Based on this observation, we aim to strengthen these alignments while addressing the remaining discrepancies to achieve more accurate parallel encoding. Our insight from Figure~\ref{fig:observation2} and Figure~\ref{fig:norm} is that \textit{KV states from independent contexts can be naturally merged into one sequence due to their similarity in direction and magnitude, attributed to the presence of an attention sink}~\citep{xiao2023efficient}. This observation reduces our challenge to addressing residual misalignments, which manifest as anomalous distributions at the initial and recent positions within each context.

Motivated by this, we propose \underline{A}daptive \underline{P}arallel \underline{E}ncoding (\textbf{APE}) to align the distribution between sequential and parallel encoding, which enables accurate and fast CAG (see Figure~\ref{fig:intro} (Bottom)). Our contributions involve:

\begin{itemize}
[itemsep=0.0pt,topsep=0pt,leftmargin=*]
\item We systematically analyze the distribution properties of attention weights in parallel encoding, focusing on the magnitude and direction of KV states across various samples and positions. Our observations identify major alignments and minor misalignments between parallel and sequential encoding for further improvement.
\item We propose APE to recover the accuracy of parallel encoding with three alignment steps: (i) Prepend a shared prefix to avoid the duplication of abnormal distribution of initial tokens. (ii) Adjust a lower attention temperature to sharpen the distribution, focusing on contextually important tokens. (iii) Apply a scaling factor to offset the increase in the magnitude of the LogSumExp value of attention scores from the context.
\item  We empirically show that (i) APE maintains 98\% and 93\% of the sequential encoding performance in RAG and ICL tasks, respectively. (ii) APE outperforms parallel encoding in RAG and ICL, yielding improvements of 3.6\% and 7.9\%, respectively. (iii) APE scales to handle hundreds of contexts in parallel, matching or exceeding sequential encoding in many-shot scenarios. (iv) APE accelerates long-context generation, achieving up to 4.5$\times$ speedup through a 28$\times$ reduction in prefilling time for a context including 128K tokens.
\end{itemize} 
\section{Background and Related Work}
\label{Preliminaries}

\subsection{Context-Augmented Generation}

This work explores CAG problems using LLMs, where user queries are enhanced with additional contexts from external databases. CAG typically involves two scenarios: RAG~\citep{asai2024reliable, gupta2024rag, gao2023retrieval}, which focuses on directly retrieving relevant information, and ICL~\citep{dong2022survey, wei2022emergent, agarwal2024many}, which emphasizes further acquiring
emergent capabilities from in-context examples.

\subsection{Parallel Encoding}
\label{sec:2.2}

Next, we present the formulation of using parallel encoding in LLMs for CAG settings. Let $\mathcal{S}$ represent the input sequence including $N$ contexts $C_1, ..., C_{N}$ and one query $Q$. Formally, this can be denoted as:
\begin{equation}
\mathcal{S} = \{\underbrace{s_{C_1, 1}, ... , s_{C_1, l_1}}_{\text{Context 1}}, \underbrace{s_{C_2, 1}, ..., s_{C_2, l_2}}_{\text{Context 2}}, ..., \underbrace{s_{C_N, 1}, ..., s_{C_N, l_N}}_{\text{Context N}}, \underbrace{s_{Q, 1}, ..., s_{Q, l}}_{\text{Query}}\}.
\end{equation}
For simplicity, we can express this as:
$\mathcal{S} = \{S_{C_1}, S_{C_2}, \ldots, S_{C_N}, S_Q\}$. Given two models $\Theta_{\text{Enc}}$ and $\Theta_{\text{Dec}}$ (which may be the same model), a response $\mathcal{R}$ is generated to the input $\mathcal{S}$ using parallel encoding in two steps:

\textbf{Pre-caching Contexts.} The first step is to encode and cache the KV states for each context independently using $\Theta_{\text{Enc}}$. For a given context $S_{C_i}$, we compute its KV states offline as $(K_{C_i}, V_{C_i}) = \Theta_{\text{Enc}}(S_{C_i})$ and store them for direct loading during inference. Specifically, we denote $K_{C_i} = \{k_{C_i, 1}, \ldots , k_{C_i, l_i}\}$ and $V_{C_i} = \{v_{C_i, 1}, \ldots , v_{C_i, l_i}\}$.

\textbf{Generating Response.} Next, the user query is augmented by all relevant pre-cached KV states to generate the response:
$\mathcal{R} = \Theta_{\text{Dec}}(S_{Q}, K_{C}, V_{C})$, where $K_{C}$, $V_{C}$ are subsets of $\{K_{C_1}, ..., K_{C_N}\}$ and $\{V_{C_1}, ..., V_{C_N}\}$, respectively.

Parallel encoding significantly improves efficiency compared to sequential encoding by reducing the complexity of prefilling from $O((l_1+...+l_N+l_Q)^2)$ (i.e., quadratic) to linear concerning the total context length. With pre-caching, the cost becomes $O((l_1+...+l_N+l_Q)\cdot l_Q)$. In the absence of pre-caching, the complexity is $O(\max(l_1^2, ..., l_N^2)+((l_1+...+l_N+l_Q)\cdot l_Q)$, which remains efficient for multiple contexts of similar length.

Prior parallel encoding approaches vary in their design of $\Theta_{\text{Enc}}$ and $\Theta_{\text{Dnc}}$. Parallel Context Windows (PCW)~\citep{ratner2022parallel} directly employs pre-trained LLMs as both, resulting in significant performance drops. Block-Attention~\citep{Sun2024BlockAttentionFE} further fine-tunes the model, successfully recovering performance in RAG tasks. Alternatively, CEPE~\citep{yen2024long} and FocusLLM~\citep{li2024focusllm} train new Transformer-based encoders using encoder-only and decoder-only architectures, respectively. These methods also differ in $\Theta_{\text{Dec}}$: CEPE trains additional cross-attention layers for processing contexts, whereas other methods directly input the context into original self-attention layers. While these trainable methods show promising results in RAG tasks, challenges remain regarding their training overheads and generalization abilities to more complex ICL scenarios. Moreover, applying parallel encoding in CAG can be viewed as a kind of memory-augmented neural networks~\citep{burtsev2020memory, de2021mention, fevry2020entities}, where external memory is directly stored into KV states.

\subsection{Attention Mechanism}

In a standard $\mathrm{Softmax}$ attention, we attend the query to all past KV states using the following formula:

\begin{equation}
    O = \mathrm{Softmax}(\frac{QK^T}{\sqrt{d}})V \quad Q\in \mathbb{R}^{n \times d} \quad K,V \in \mathbb{R}^{m \times d},
    \label{eq:attention}
\end{equation}

where $Q$ is the query state, and $K$ and $V$ denote the key and value states, respectively. Previous research has revealed several significant insights into the distribution of attention weights (i.e., $\mathrm{Softmax}(\frac{QK^T}{\sqrt{d}})$).

\textbf{Attention Sink.} StreamingLLM~\citep{xiao2023efficient} identifies the presence of an ``attention sink'' in LLMs, a token that receives a significantly higher attention score than other tokens but provides limited semantic information. It observes that the attention sink exists in the initial token and influences the following tokens.

\textbf{Position Embedding.} To effectively process sequential input, LLMs require position embeddings, such as absolute position embeddings~\citep{vaswani2017attention, devlin2018bert} and relative position embeddings~\citep{su2024roformer, press2021train}. However, the introduction of position embedding not only limits the context window to the training length~\citep{chen2023extending} but also results in the ``lost in the middle"~\citep{liu2024lost} issue, where LLMs struggle to produce correct answers when relevant information locates in the middle of the context.

\section{Observations}
\label{obs}

\begin{figure}[!h]
    \centering
    \begin{subfigure}[b]{0.46\textwidth}
        \centering
        \includegraphics[width=\textwidth]{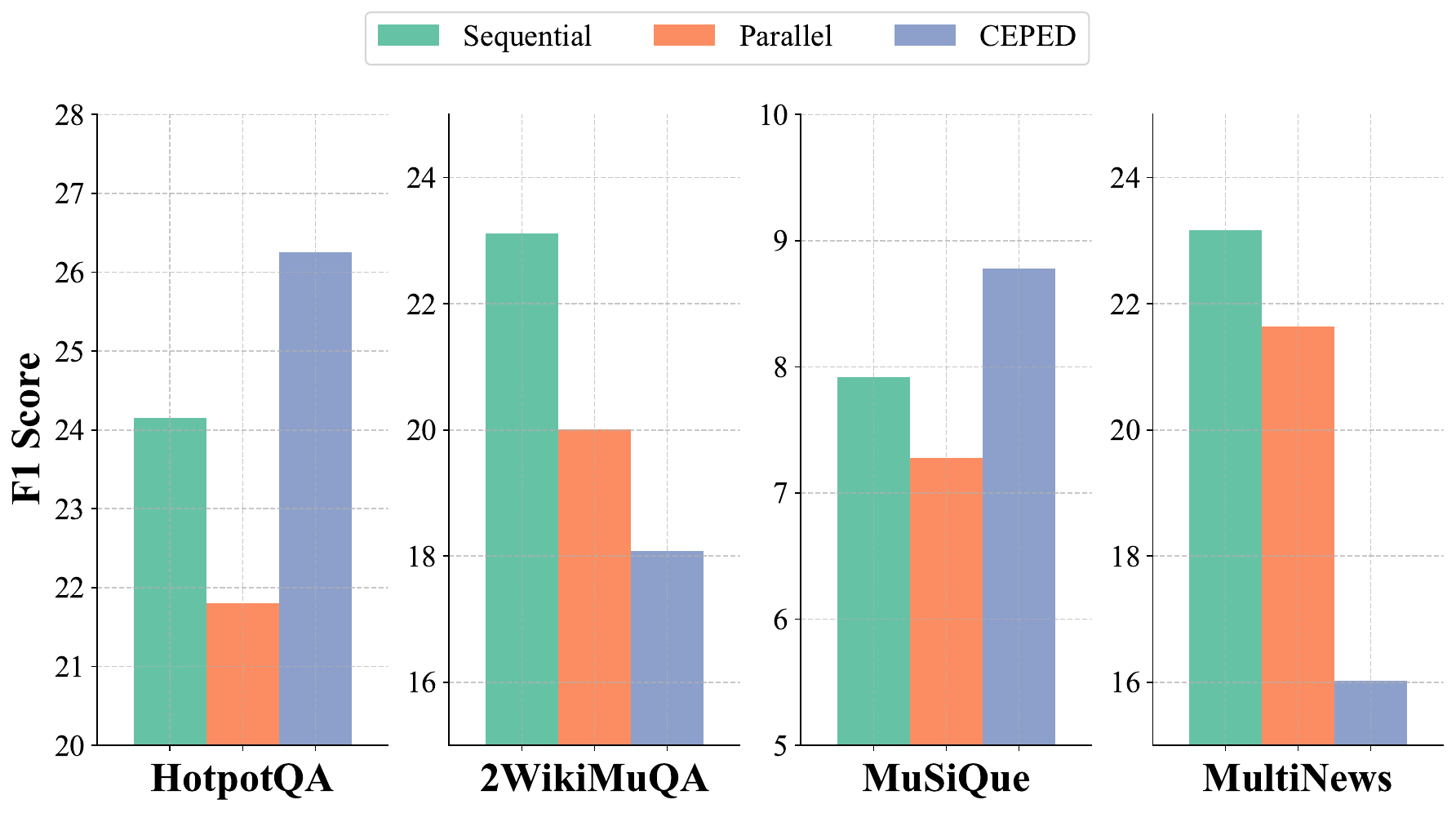}
        \caption{Retrieval-augmented Generation}
        \label{fig:obseravtion1rag}
    \end{subfigure}
    \hfill
    \begin{subfigure}[b]{0.52\textwidth}
        \centering
        \includegraphics[width=\textwidth]{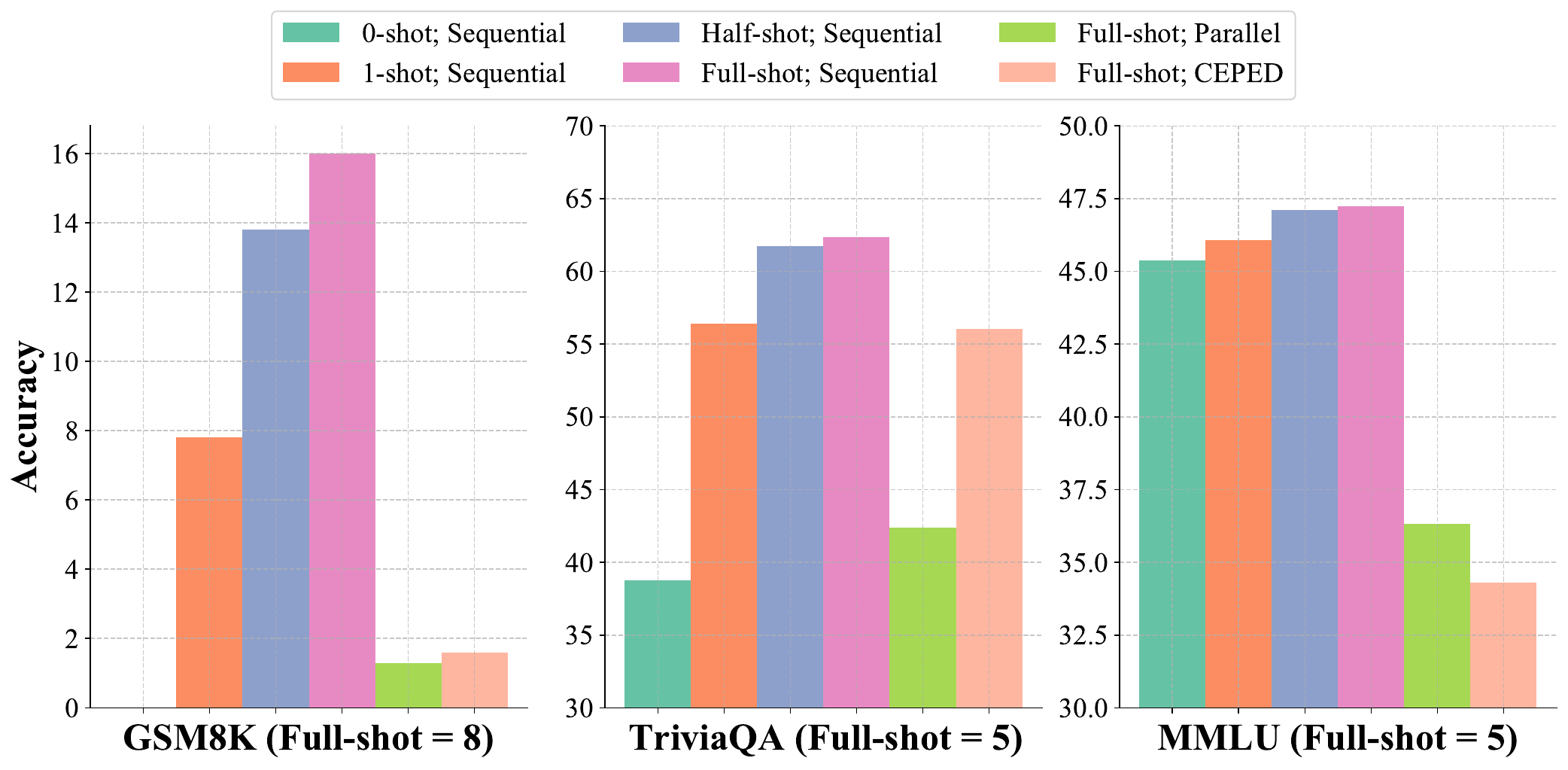}
        \caption{In-context Learning}
        \label{fig:obseravtion1icl}
    \end{subfigure}
    \caption{\textbf{Comparison of sequential encoding, parallel encoding, and CEPED in RAG and ICL scenarios.} Parallel encoding and CEPED degrades performance, especially on tasks such as GSM8K that requires reasoning ability.}
    \label{fig:obseravtion1}
\end{figure}

In Section~\ref{obs2}, we evaluate sequential encoding, parallel encoding, and CEPE-Distilled (CEPED)~\citep{yen2024long} using the \textsc{LLaMA-2-7B-chat} model\footnote{We use the \textsc{LLaMA-2} model for CEPED, as it is the only supported model. For other analyses, we employ \textsc{LLaMA-3}.}. Figure~\ref{fig:obseravtion1} presents our findings on various RAG and ICL tasks, highlighting the limitations of trainable approaches in generalizing to complex reasoning tasks. Next, we explore the alignments and misalignments between parallel encoding and sequential encoding in Section~\ref{obs1}, providing insights into why parallel encoding remains effective and identifying opportunities for further improvement.

\subsection{Trainable Approaches are only Effective for Easy Tasks.}
\label{obs2}
In Figure~\ref{fig:obseravtion1}, we compare the performance of different context encoding methods on RAG and ICL tasks, with detailed setups described in Appendix~\ref{app:obs1}. Our analysis of the long-context RAG capability on LongBench~\citep{bai2023longbench} is showcased in Figure~\ref{fig:obseravtion1rag}. Despite accessing more passages, CEPED only surpasses the sequential baseline in two of the three QA tasks, and it even notably underperforms parallel encoding in the summarization task (MultiNews), which requires synthesizing information from the entire context. We hypothesize that CEPED cannot process complex tasks since the encoder and decoder are only trained on the unlabeled pre-training corpus without instruction-tuning on high-quality QA samples. This conclusion is further supported by the results of ICL tasks (see Figure~\ref{fig:obseravtion1icl}), where CEPED performs on par with the 1-shot sequential encoding baseline on TriviaQA but falls short of it on GSM8K and MMLU, despite using much more examples. The latter involves reasoning steps that are hard for the ill-trained model to understand. In conclusion, fine-tuning models to improve parallel encoding requires (i) more diverse and labeled data and (ii) resource-intensive instruction-tuning (e.g., SFT or RLHF~\citep{ouyang2022training}). Given this unfavorable trade-off between training costs and model \mbox{capabilities, we propose developing a training-free method to improve the performance of parallel encoding.}

\subsection{Comparing Parallel Encoding and Sequential Encoding.}
\label{obs1}
\begin{figure}[t]
    \centering
    \includegraphics[width=\textwidth]{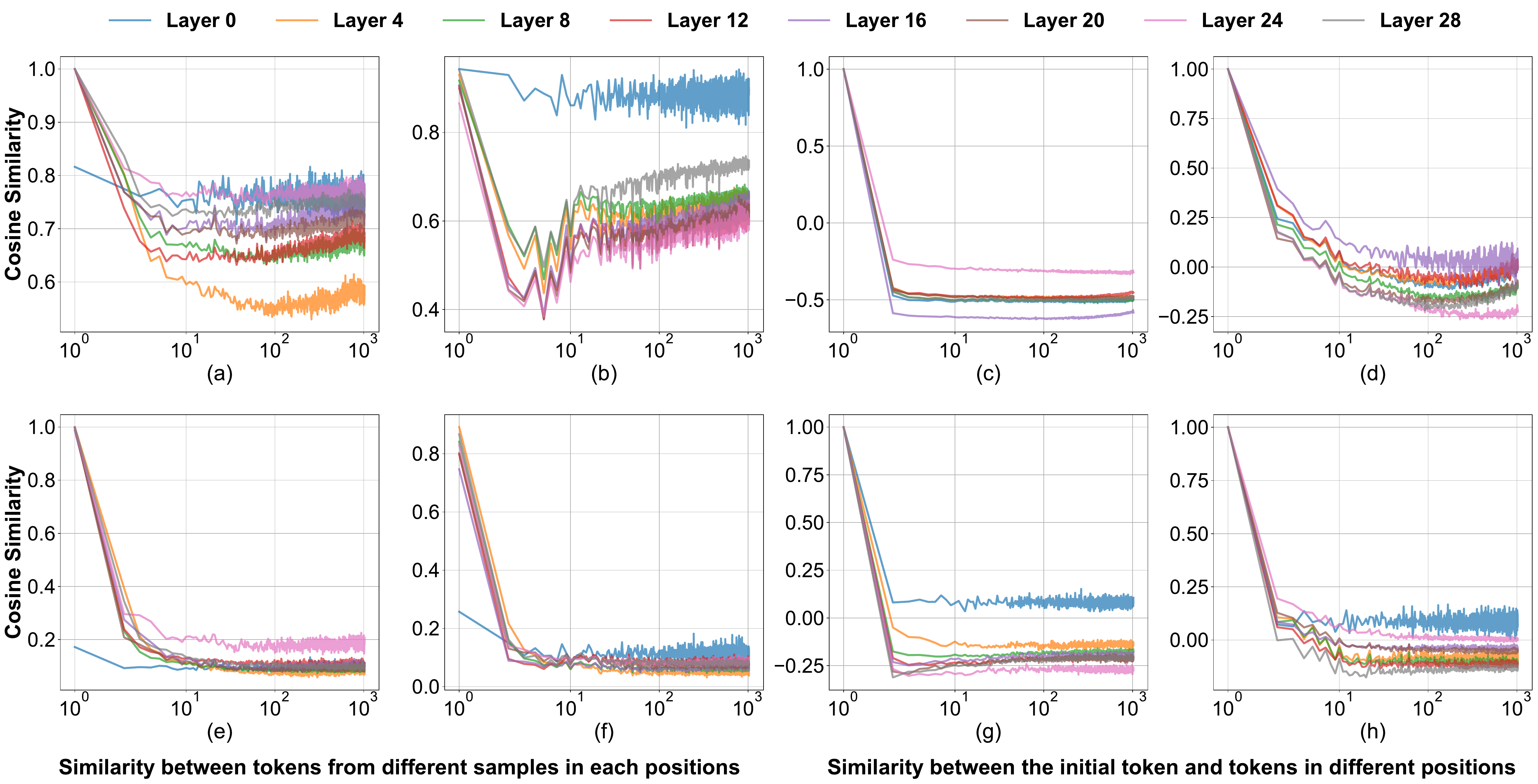}
    \caption{\textbf{Top Left:} 
 Both \textsc{LLaMA-3-8B-Instruct} (a) and \textsc{Mistral-7B-Instruct-v0.3} (b) exhibit a cosine similarity larger than 0.9 for the key states from distinct initial tokens. \textbf{Top Right:} Initial token's key states show similar negative values to those from other positions for \textsc{LLaMA-3-8B-Instruct} (c) and \textsc{Mistral-7B-Instruct-v0.3} (d) models. \textbf{Bottom:} Value states exhibit patterns similar to those observed in key states. The X-axis shows the positions of key and value states on a logarithmic scale. Visualizations and analyses for more base models are provided in Appendix~\ref{app:obs2}.}
    \label{fig:observation2}
\end{figure}

\begin{figure}
    \centering
    \begin{subfigure}[b]{0.24\textwidth}
        \centering
        \includegraphics[width=\textwidth]{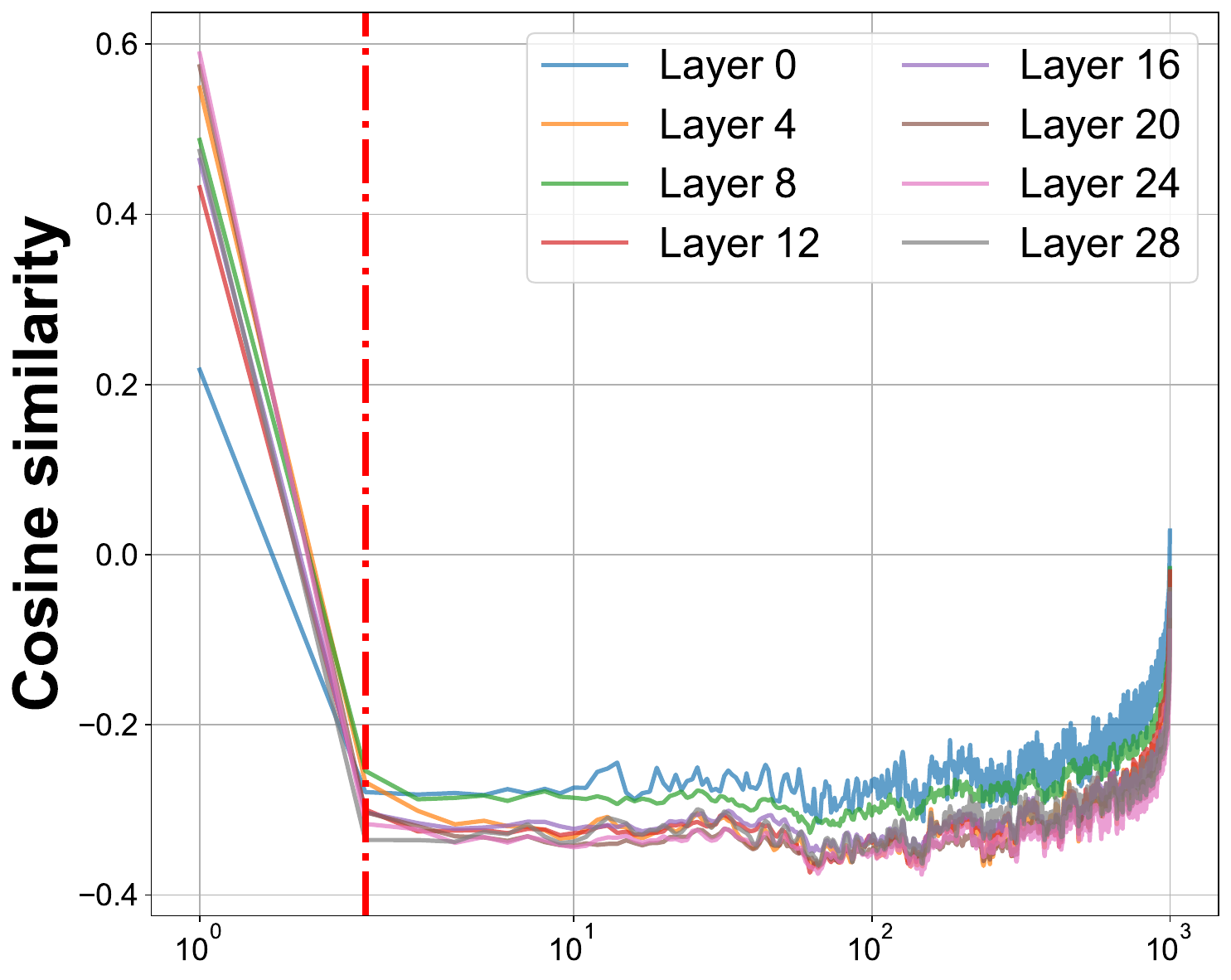}
        \caption{Query-Key Similarity}
        \label{fig:qk_sim}
    \end{subfigure}
    \hfill
    \begin{subfigure}[b]{0.24\textwidth}
        \centering
        \includegraphics[width=\textwidth]{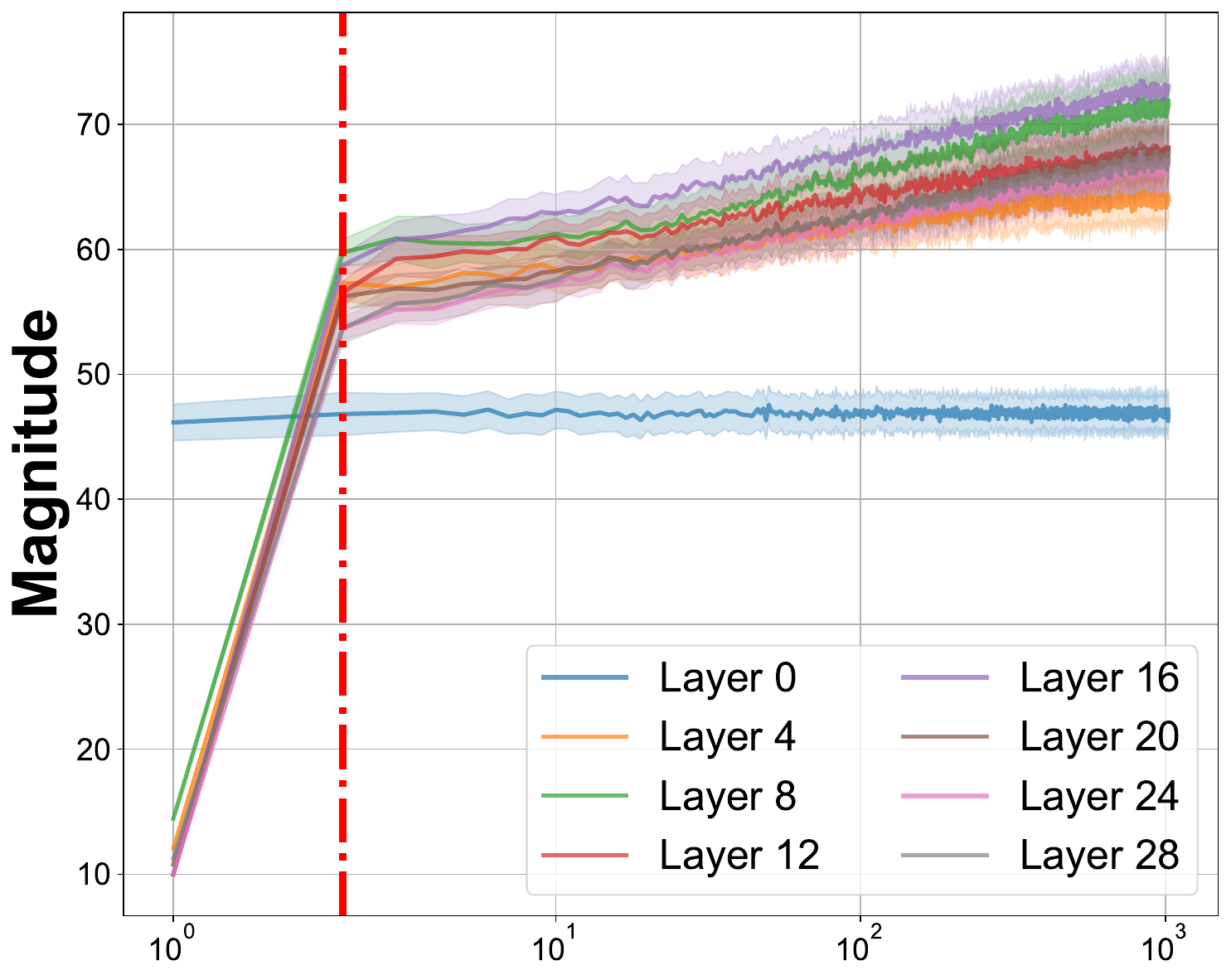}
        \caption{Key Magnitude}
        \label{fig:norm_k}
    \end{subfigure}
    \hfill
    \begin{subfigure}[b]{0.24\textwidth}
        \centering
        \includegraphics[width=\textwidth]{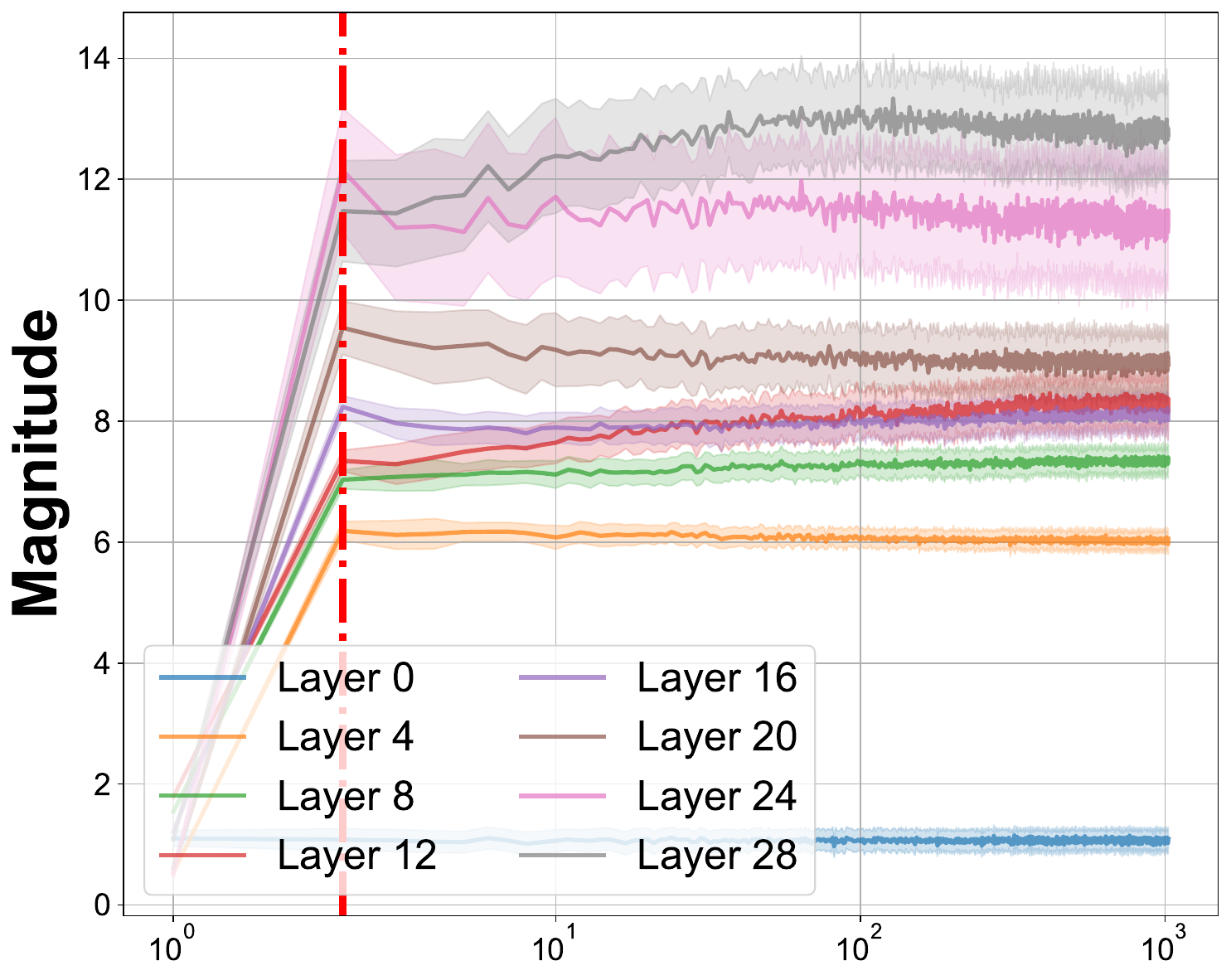}
        \caption{Value Magnitude}
        \label{fig:norm_v}
    \end{subfigure}
    \hfill
    \begin{subfigure}[b]{0.24\textwidth}
        \centering
        \includegraphics[width=\textwidth]{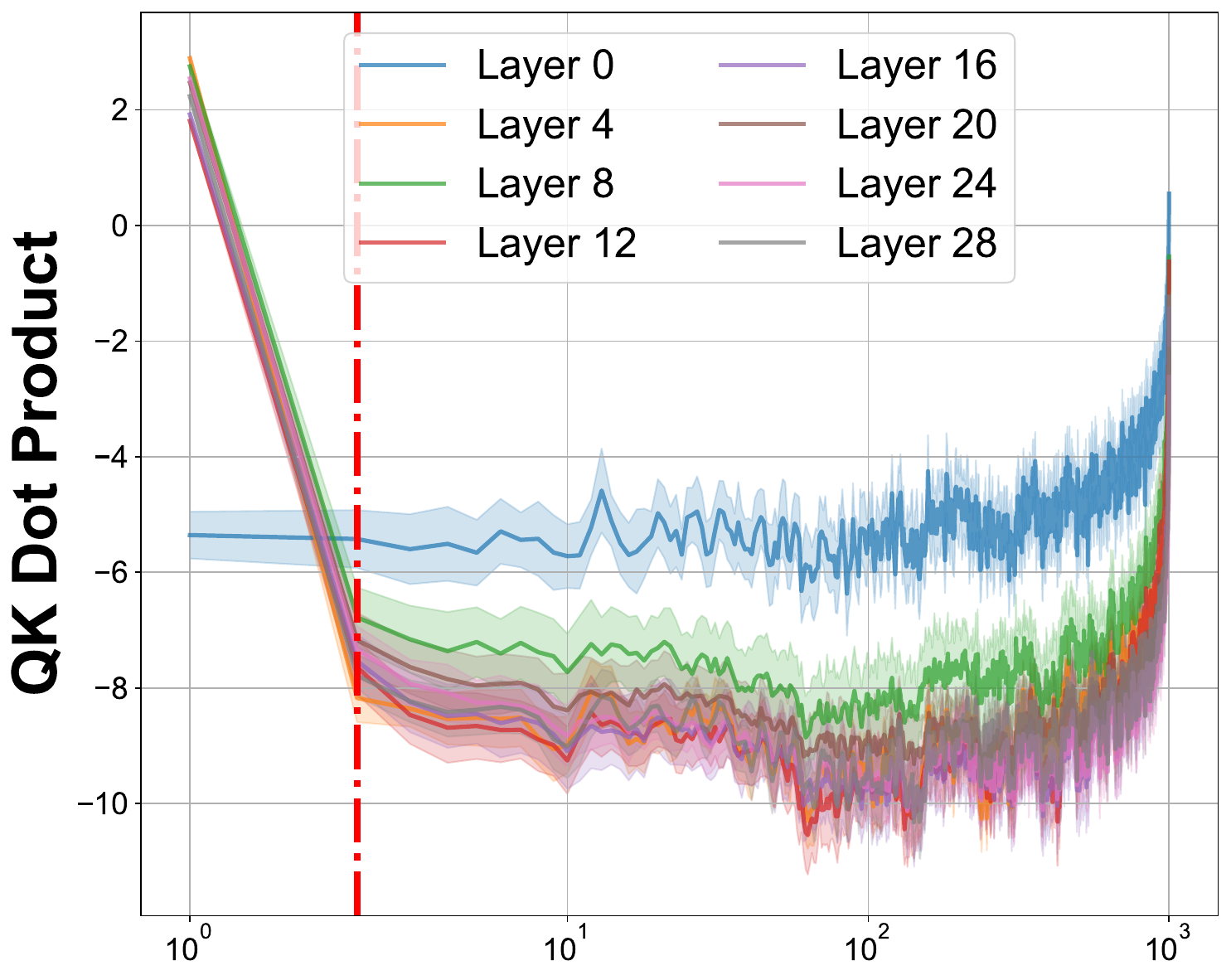}
        \caption{Query-Key Product}
        \label{fig:attn_score}
    \end{subfigure}
    \caption{\textbf{Visualization of Different Components in Attention.} \textbf{(a)} 
The cosine similarity between query and key states increases as the distance between their positions decreases. \textbf{(b)} The magnitudes of key states show a slowly upward trend as position increases. \textbf{(c)} The magnitude of value states remain constant across positions. \textbf{(d)} Query-key dot products keep consistently low values except at initial and recent positions. A red dashed line marks the anomalous region for the first two tokens in all figures. The X-axis shows positions of KV states on a log scale. Results are measured with the \textsc{LLaMA-3-8B-Instruct} model. Visualizations and analyses for more base models are provided in Appendix~\ref{app:obs2}.}
    \label{fig:norm}
\end{figure}

In Figure~\ref{fig:obseravtion1}, we observe that parallel encoding still holds promise, as it can generate reasonable responses without further modifications. This finding is non-trivial as contexts are encoded into KV states separately without guarantee that these states can be compared or combined. However, our analysis reveals that the attention mechanism naturally builds alignments between KV states from different positions in independent contexts similar to sequential encoding. To clarify this, Figure~\ref{fig:observation2} focuses on the impact of the attention sink~\citep{xiao2023efficient}, where we visualize the direction of KV states for different samples and positions. In Figure~\ref{fig:norm}, we further visualize the distribution of various components in the $\mathrm{Softmax}$ attention, resulting in several findings.

\begin{wrapfigure}{r}{0.38\textwidth}
    \centering
    \captionsetup{justification=centering}
    \vspace{-1.5em}
    \includegraphics[width=0.37\textwidth]{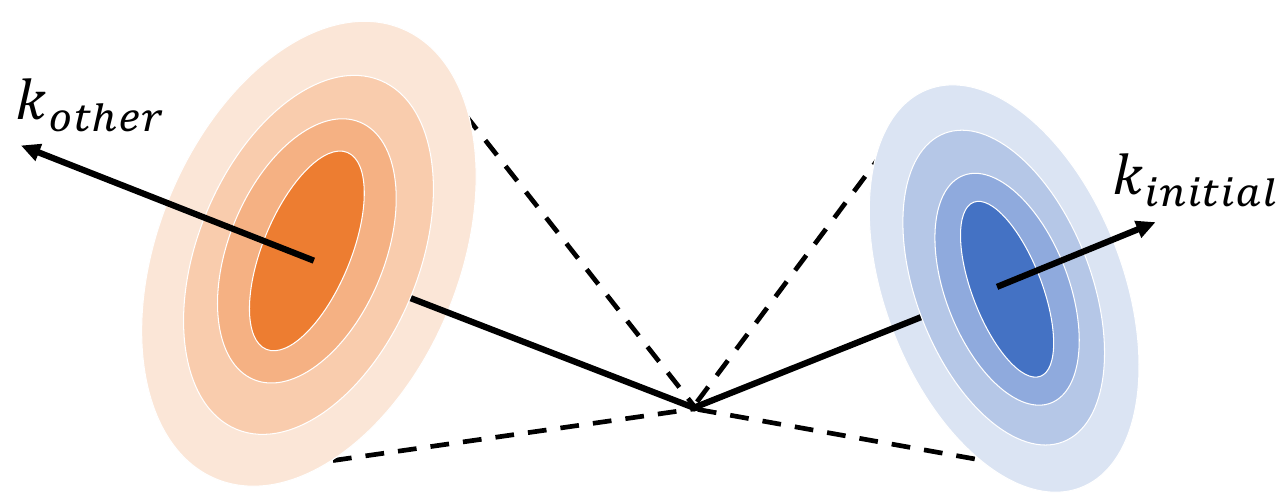}
    \caption{\textbf{Geometry of Key States.}}
    \label{fig:illsuration}
    \vspace{-1.5em}
\end{wrapfigure}

\textbf{Key states from different contexts are similar.}  
In Figure~\ref{fig:observation2}a and \ref{fig:observation2}b, we measure the cosine similarity between the key states of different initial tokens for the \textsc{LLaMA-3-8B-Instruct} and \textsc{Mistral-7B-Instruct-v0.3} models, which consistently yields a value close to 1. This observation indicates that the direction of the initial key state remains invariant mainly across different inputs. Figure \ref{fig:observation2}c and \ref{fig:observation2}d further analyze the similarity between the initial key states and their subsequent states, where we observe comparable negative values from different positions. Therefore, the angles between the initial key states and their subsequent states are similar and significantly larger than the angles between different initial key states, as demonstrated in Figure~\ref{fig:illsuration}. It suggests that the direction of key states remains relatively consistent across contexts, as they are primarily decided by the initial key states, which exhibit similar directions across examples. These findings, combined with the small variance in key state magnitudes across examples in Figure~\ref{fig:norm_k}, indicate that key states from different contexts share similar directions and magnitudes, making them comparable. 
 
\begin{wrapfigure}{r}{0.45\textwidth}
    \centering
    \vspace{-1.3em}
    \captionsetup{justification=centering}
    \includegraphics[width=0.44\textwidth]{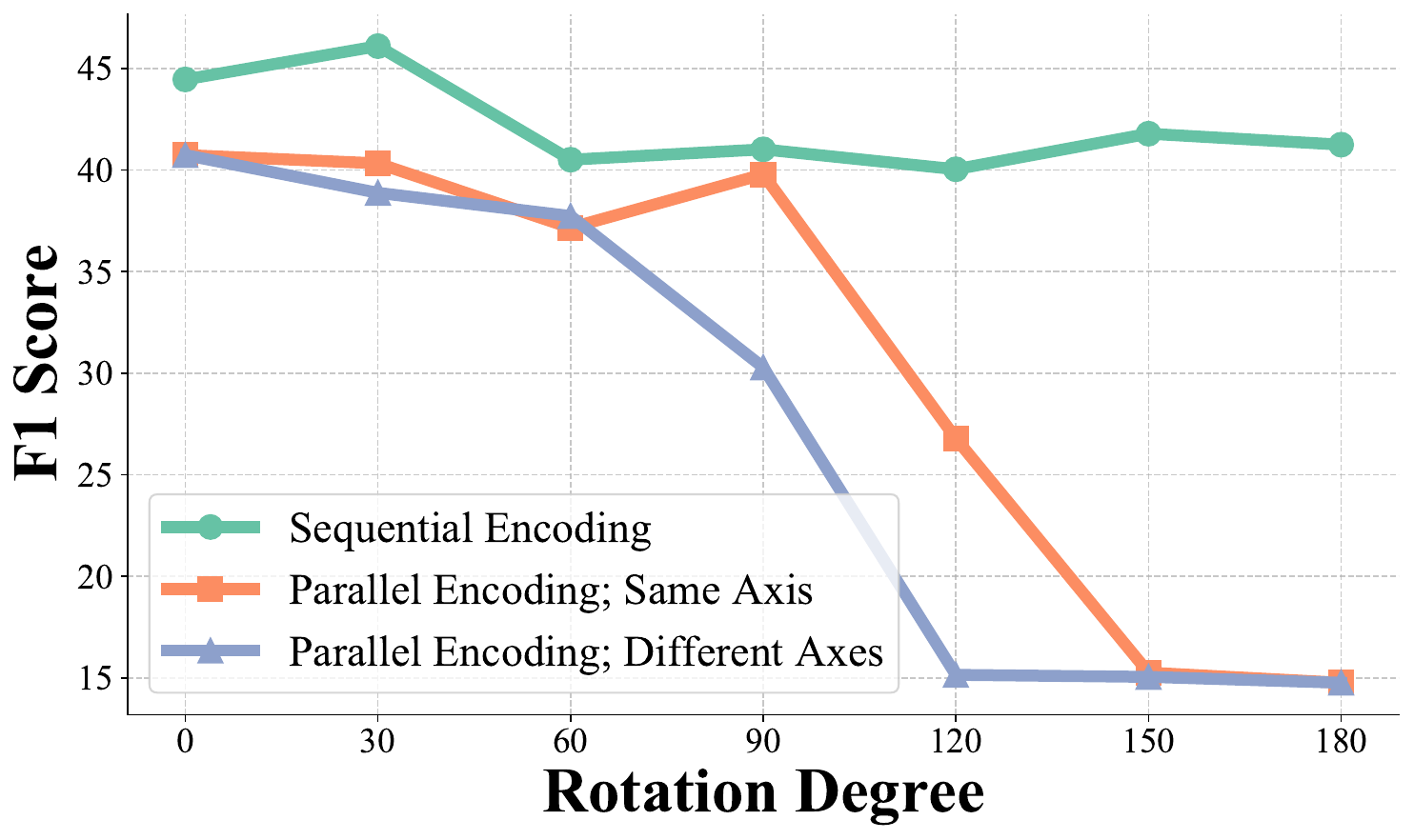}
    \caption{\textbf{Rotation Analysis on the First Token}}
    \vspace{-0.6em}
    \label{fig:rotation}
\end{wrapfigure}

To further understand this, we experiment on HotPotQA using the \textsc{LLaMA-3-8B-Instruct} model. Our analysis involves applying rotations of varying degrees around random axes to the initial key states. For parallel encoding, we explore two rotation modes: one using the same rotation axis for all contexts and another employing a random rotation axis for each context. Figure~\ref{fig:rotation} reveals that sequential encoding keeps performance across various rotation degrees. In contrast, both modes in parallel encoding deteriorate when rotations exceed 150 degrees. This effect arises from the duplication of initial key states, intensifying our rotations' impact. Notably, using separate axes for each context leads to an earlier breakdown beginning at 90 degrees. This mode disrupts the directional similarity of key states with different initial tokens (i.e., $k_{\text{initial}}$) in Figure~\ref{fig:illsuration} and enlarges the angle between key states from different contexts.

\textbf{Values states from different contexts can be combined.} In Equation \eqref{eq:attention}, all value states are combined through a weighted summation, where the $\mathrm{Softmax}$ operator would normalize the weights of all value states to sum to 1. This normalization indicates that the magnitude of current value states is determined solely by those from previous positions, resulting in a similar $L^2$ norm across positions, as shown in Figure~\ref{fig:norm_v}. Additionally, the small variance shows that the magnitudes are comparable among samples. This finding, coupled with a similar direction across samples and positions in Figure~\ref{fig:observation2} (Bottom), indicates the possibility of combining value states.

\textbf{Opportunities for improvement.} Despite the KV states exhibiting similarity across contexts for most positions, the residual misalignments in Figure~\ref{fig:norm} still severely reduce accuracy. We summarize them as follows:
\begin{itemize}
[itemsep=0.0pt,topsep=0pt,leftmargin=*]
\item In Figure \ref{fig:norm}, we observe a notable discrepancy in direction and magnitude for the initial positions, leading to large QK dot products at these positions in Figure \ref{fig:attn_score}. They are identified as an anomaly in the context.
\item Figure~\ref{fig:attn_score} shows the dot products between the query state and all past key states, revealing a notable increase when the states are positioned close to each other, as reflected in the larger similarity observed in Figure~\ref{fig:qk_sim}.
\end{itemize}
\section{Adaptive Parallel Encoding}
With all the lessons learned in Section~\ref{obs}, we will design our \textbf{APE} to address the residual misalignments. APE enables a seamless shift to parallel encoding without requiring training while maintaining most of the model's capabilities. Our approach adaptively aligns the distribution of attention weights between sequential and parallel encoding via three steps as illustrated in Figure~\ref{fig:intro}, thereby boosting efficiency and performance. 

\subsection{Prepending Shared Prefix.}
\label{sec:shared_prefix}
Figure~\ref{fig:norm} shows that the distribution of the first few tokens differs significantly from that of subsequent tokens. This discrepancy poses a challenge when encoding contexts in parallel due to duplicating these abnormal KV states. To address this issue, we propose a simple yet effective solution: prepending a shared prefix to all contexts. This approach ensures that these KV states appear only once in each generation step. In practice, the choice of prefix varies with the model and task. We use existing system prompts and instructions as the shared prefix when available. Otherwise, we will insert a few newline characters (i.e., ``\textbackslash n") before all contexts.

Although the later positions are not identified as abnormal, we still observe instability at the start of LLM inputs. To mitigate this issue, we may also consider extending the existing prefix with more newline characters.

\begin{figure}
    \centering
    \begin{subfigure}[b]{0.24\textwidth}
        \centering
        \includegraphics[width=\textwidth]{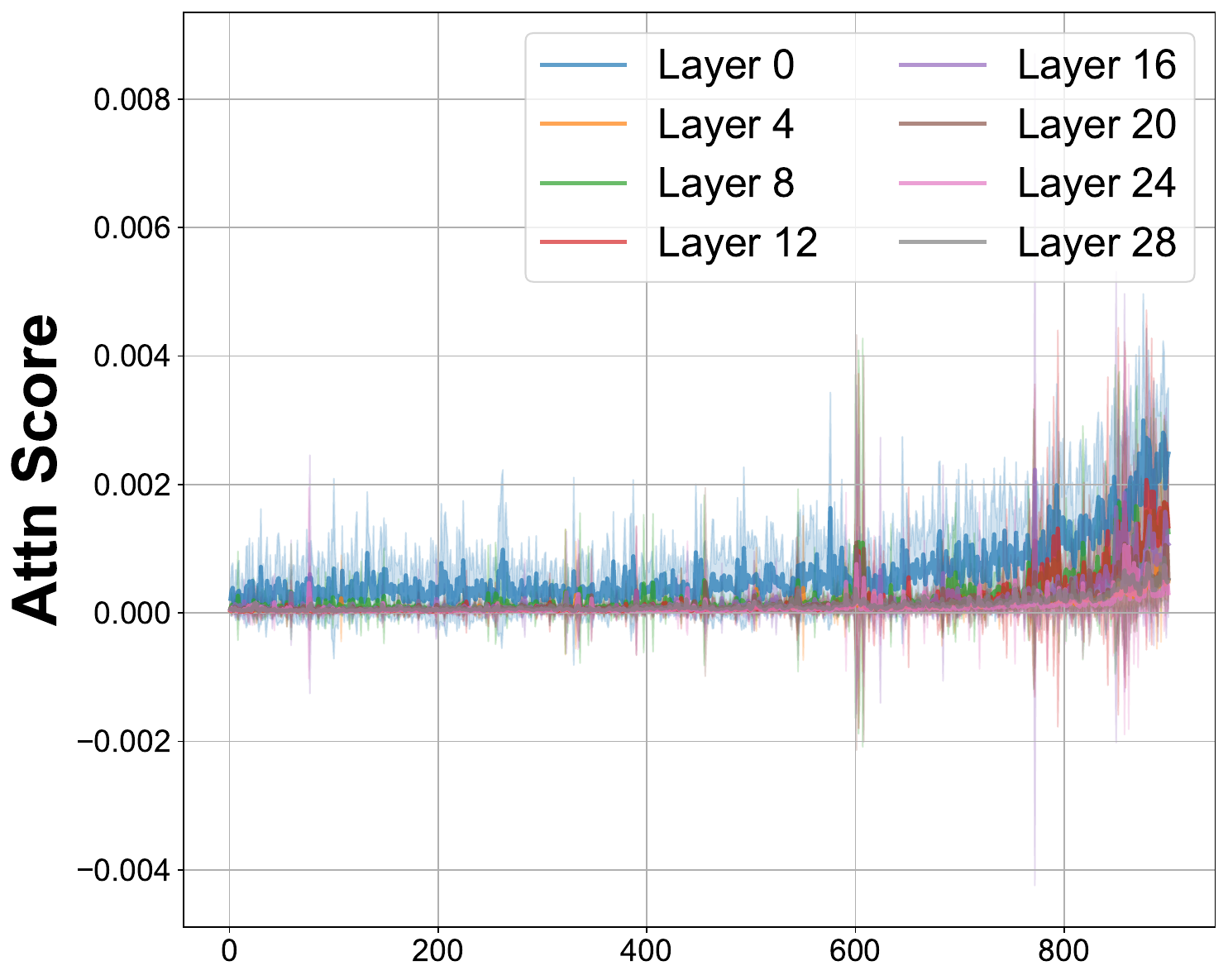}
        \caption{Sequential}
        \label{fig:attn_score_s}
    \end{subfigure}
    \hfill
    \begin{subfigure}[b]{0.24\textwidth}
        \centering
        \includegraphics[width=\textwidth]{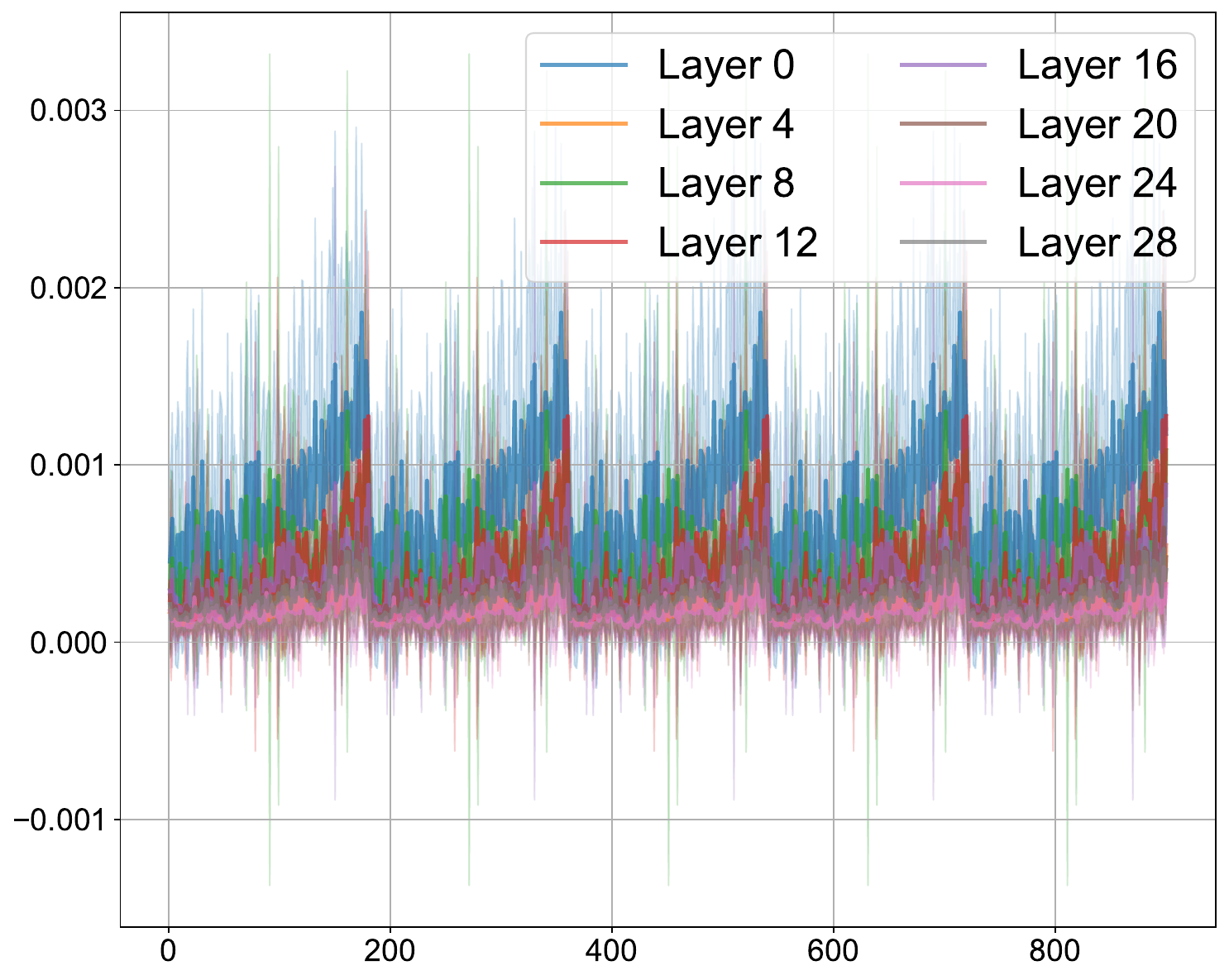}
        \caption{Parallel  (T = 1.0)}
        \label{fig:attn_score_p}
    \end{subfigure}
    \hfill
    \begin{subfigure}[b]{0.24\textwidth}
        \centering
        \includegraphics[width=\textwidth]{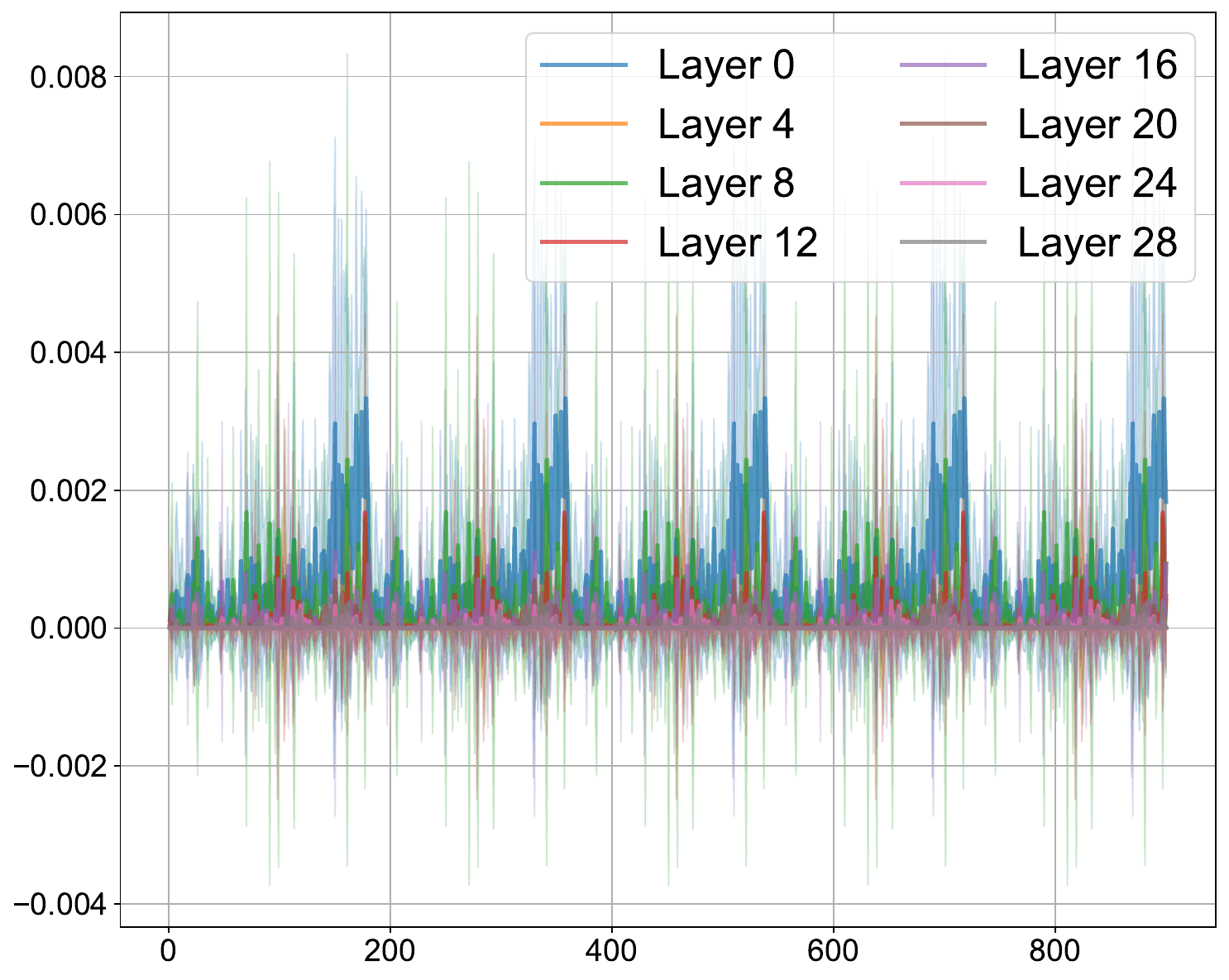}
        \caption{Parallel (T = 0.2)}
        \label{fig:attn_score_p2}
    \end{subfigure}
    \hfill
    \begin{subfigure}[b]{0.24\textwidth}
        \centering
        \includegraphics[width=\textwidth]{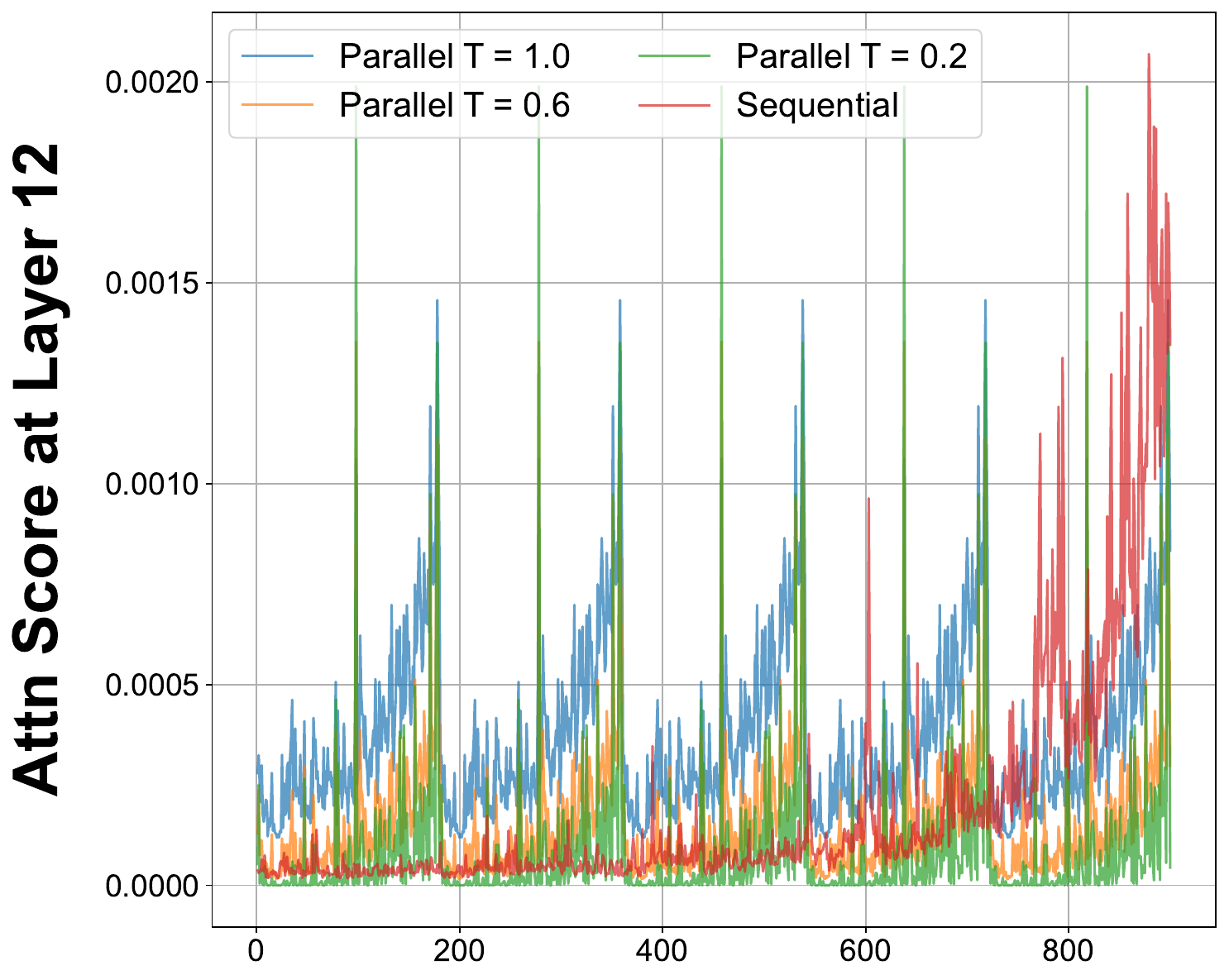}
        \caption{Parallel vs. Sequential}
        \label{fig:attn_score_p3}
    \end{subfigure}
    \caption{\textbf{Comparison of Attention Weight Distribution within Contexts.} \textbf{(a)} 
Sequential encoding allocates high attention scores to neighboring tokens. \textbf{(b)} Parallel encoding distributes attention scores more uniform across neighboring tokens from all contexts. \textbf{(c)} Adjusting the temperature $T$ sparsifies the distribution. \textbf{(d)} After adjustment, the distribution in parallel encoding becomes similar to sequential encoding. The X-axis represents token positions.} 
    \label{fig:alignment}
\end{figure}

\subsection{Adjusting Attention Temperature.}

In Figure \ref{fig:attn_score}, the value of QK dot products increases as the relative distance decreases, with a notably sharper rise when the distance approaches zero. To show its impact on parallel encoding, we set a 50-token prefix and query, encoding the remaining 900 tokens either sequentially or in five parallel chunks, with attention distributions shown in Figure~\ref{fig:alignment}. 
Comparing Figure~\ref{fig:attn_score_p} with \ref{fig:attn_score_s}, duplicating neighboring KV states in parallel encoding will disperse the query's attention to multiple contexts, resulting in a more uniform attention distribution. We adjust the attention temperature $T$ to a value less than 1 to refocus on the most relevant tokens, sharpening the distribution after the $\mathrm{Softmax}$ operation. The comparison between different $T$ is shown in Figure \ref{fig:attn_score_p2} and \ref{fig:attn_score_p3}.

\subsection{Adding Scaling Factor.}

\begin{wrapfigure}{r}{0.32\textwidth}
    \centering
    \captionsetup{justification=centering}
    \vspace{-1em}
    \includegraphics[width=0.31\textwidth]{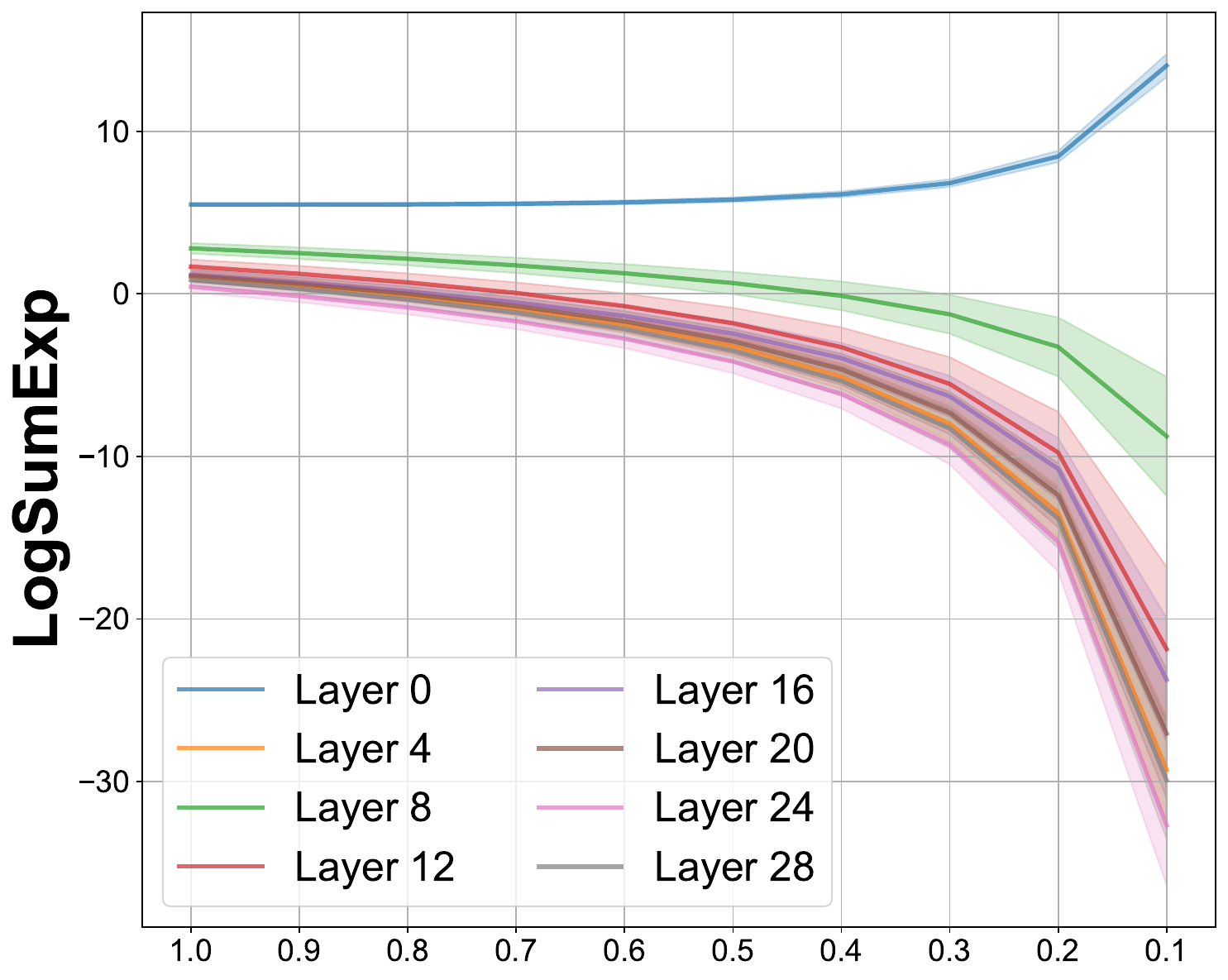}
    \caption{\textbf{Parallel w/ Different $T$.}}
    \label{fig:alignment2}
    \vspace{-1.5em}
\end{wrapfigure}

While adjusting the temperature sharpens the attention distribution among context tokens, it will also alter the overall attention allocated to the whole context, as indicated by the LogSumExp value in Figure~\ref{fig:alignment2}. Specifically, when the sum of the original QK dot product values in a given layer is significantly greater than 0, reducing temperature amplifies these positive values, resulting in an increased, positive LogSumExp value. Conversely, when the sum is closer to 0, lowering temperature has a stronger effect on the negative QK dot products, leading to a decreased, negative LogSumExp value. These effects generally increase the absolute value of LogSumExp(QK). To compensate for these changes, we introduce a scaling factor $S < 1$ to reduce this absolute value.





\subsection{Formulation.}

Given these three steps, we can formulate the modified attention in APE. We begin with the standard $\mathrm{Softmax}$ attention, where $Q$, $K$, and $V$ are the query, key, and value states, respectively. We use the subscript ${C_i}$ for elements from the context ${C_i}$, while those without a subscript correspond to user queries or generated texts.

\begin{align}
O &= \mathrm{Softmax}\left(\frac{Q[K_{C_1}^\top, \ldots, K_{C_N}^\top, K^\top]}{\sqrt{d}}\right) \times [V_{C_1}, \ldots, V_{C_N}, V] \\
&= \frac{ [A_{C_1}, \ldots, A_{C_N}, A]}{\sum_{i=1}^{N}\sum_{j=1}^{l_{C_i}}a_{C_{i}, j}+ \sum_{j=1}^{l}a_{j}}\times [V_{C_1}, \ldots, V_{C_N}, V], \\
\text{where }A_{C_i} &= [\exp \frac{Qk_{C_i, 1}^\top}{\sqrt{d}}, \ldots, \exp \frac{Qk_{C_i, l_{C_i}}^\top}{\sqrt{d}}]\text{ and }a_{C_{i}, j} = \exp\frac{Qk_{C_i, j}^\top}{\sqrt{d}}.\text{ Similar for }A\text{ and }a_{j}.\notag
\end{align}

After incorporating the proposed changes, the formula for our refined attention calculation becomes:

\begin{align}
\label{algo:ape}
    O' &= \frac{[A_{P}, A'_{C_1}, \ldots, A'_{C_N}, A]}{\sum_{j=1}^{l_P}a_{P, j} + (\sum_{i=1}^{N}\sum_{j=1}^{l_{C_i}}a'_{C_i, j})^{S}+ \sum_{j=1}^{l}a_{j}}\times [V_P, V_{C_1}, \ldots, V_{C_{N}}, V], \\
    &\text{where }A'_{C_i} = [ \exp \frac{Qk_{C_i, 1}^\top}{T\sqrt{d}}, \ldots,  \exp \frac{Qk_{C_i, l_{C_i}}^\top}{T\sqrt{d}}] \cdot (\sum_{i=1}^{N}\sum_{j=1}^{l_{C_i}}a'_{C_i, j})^{S-1}\text{ and }a'_{C_{i}, j} = \exp\frac{Qk_{C_i, j}^\top}{T\sqrt{d}}.\notag
\end{align}

$A_P$ represents the attention weights for the shared prefix while $A$ denotes that for query and generated tokens. The attention temperature $T$ and the scaling factor $S$ for the context are less than 1. Appendix~\ref{app:algorithm} provides a detailed deduction of this formula for better understanding. All these modifications are compatible with fast attention implementations such as flash attention~\citep{dao2022flashattention} by computing the context and non-context KV states separately and merging them into the attention output. This process only incur a negligible overhead.

For the choice of hyperparameters, we conduct a greedy search over a small validation set. If no prefix is provided, we begin by adding two ``$\backslash$n" and increase the prefix length by 10, 20, and 40. $S$ and $T$ are searched in the ranges [0.1, 1.0] using 0.1 step sizes. We use $S\cdot T$ instead of $S$ as the scaling factor to simplify our search.
\section{Experiments}

Empirically, we present the effectiveness and efficiency of APE in CAG scenarios such as RAG and ICL. Since we focus on context encoding problems, we do not include comparisons with long-context LLMs. Specifically,
\begin{itemize}[itemsep=0.0ex,topsep=0pt,leftmargin=*]
	\item In Section~\ref{sec:longbench}, APE can maintain 98\% of the accuracy on ChatRAG-Bench compared to sequential encoding. Furthermore, it improves 3.3\% performance for RAG on LongBench by retrieving more and longer contexts. 
	\item In Section~\ref{icl}, APE outperforms parallel encoding by 7.9\% on average in three ICL tasks. Moreover, APE can maintain 93\% of the accuracy achieved by sequential encoding when using the same number of examples.
        \item In Section~\ref{sec:loft}, APE can scale to many-shot CAG tasks, effectively encoding hundreds of texts in parallel.
	\item In Section~\ref{sec:efficiency}, APE achieves 4.5$\times$ faster inference for 128k context through 28$\times$ reduction in prefilling time.
\end{itemize}

\subsection{Retrieval-Augmented Generation.}
\label{sec:longbench}

In the context of RAG tasks, we validate that APE retains most of the sequential encoding capability while accommodating more and longer contexts, mitigating retrieval errors, and outperforming encoding baselines.


\subsubsection{Retrieval for Multi-turn Question Answering.}

\textbf{Setup.} APE is evaluated on five conversational QA tasks using ChatRAGBench~\citep{liu2024chatqa}. For each query, we prepare about 100 text chunks. Three retrievers of varying quality are employed to retrieve up to the top-5 chunks for evaluation, including Contriever~\citep{izacard2021unsupervised}, GTE-base~\cite{li2023towards}, and Dragon-multiturn~\cite{liu2024chatqa}. We use \textsc{Llama3-ChatQA-1.5-8B} as the base model. To fairly measure performance drop after our modifications, the same retrieved texts are used for APE and sequential encoding.

\textbf{Results.} Table~\ref{tab:chatragbench} shows that switching from sequential encoding to APE results in performance drops of 0.51\%, 0.92\%, and 1.14\% across different retrievers, respectively. While this drop increases with retriever quality, APE still keeps 97\% of the sequential encoding performance for the best retriever. By increasing the text chunk length for $5$ times, APE directly inputs all texts without any retrieval process, achieving superior performance.

\begin{table}[t]
\centering
\caption{\textbf{Comparison between APE and sequential encoding using three retrievers on ChatRAG-Bench.}}
\resizebox{\linewidth}{!}{\setlength{\tabcolsep}{3mm}{
\begin{tabular}{l|cccccc}
\toprule
Method & INSCIT & Doc2Dial & TopicCQA & Qrecc & QuAC & Average \\
\midrule
Contriever, Sequential & 19.97 & 23.85 & 30.49 & 46.75 & 26.57 & 29.53 \\
Contriever, APE & 19.88 & 23.28 & 28.84 & 46.28 & 26.80 & 29.02 \\
$\Delta$ & \textcolor{blue}{-0.09} & \textcolor{blue}{-0.57} & \textcolor{blue}{-1.65} & \textcolor{blue}{-0.47} & \textcolor{red}{+0.23} & \textcolor{blue}{-0.51} \\
\midrule
GTE-base, Sequential & 21.58 & 32.35 & 33.41 & 46.54 & 30.69 & 32.91 \\
GTE-base, APE & 20.85 & 30.99 & 31.92 & 45.83 & 30.35 & 31.99 \\
$\Delta$ & \textcolor{blue}{-0.73} & \textcolor{blue}{-1.36} & \textcolor{blue}{-1.49} & \textcolor{blue}{-0.71} & \textcolor{blue}{-0.34} & \textcolor{blue}{-0.92} \\
\midrule
Dragon-multiturn, Sequential & 25.42 & 36.27 & 36.10 & 49.01 & 35.12 & 36.38 \\
Dragon-multiturn, APE & 23.84 & 34.93 & 33.80 & 48.70 & 34.92 & 35.24 \\
$\Delta$ & \textcolor{blue}{-1.58} & \textcolor{blue}{-1.34} & \textcolor{blue}{-2.30} & \textcolor{blue}{-0.31} & \textcolor{blue}{-0.20} & \textcolor{blue}{-1.14} \\
\midrule
All texts, APE & \textbf{27.22} & 36.13 & 35.72 & \textbf{49.15} & \textbf{35.70} & \textbf{36.78} \\
\bottomrule
\end{tabular}}}
\label{tab:chatragbench}
\end{table}

\subsubsection{Retrieval for Long-context Understanding.}

\textbf{Setup.} Our evaluation involves eight tasks on LongBench~\citep{bai2023longbench}. Given the long context,  we split it into chunks with a size of $M$ words, employ Contriever~\citep{izacard2021unsupervised} to compute the embeddings of all
chunks and the query and retrieve the top-$N$ chunks according to the cosine similarity of
their embeddings to the query embedding. $M$ and $N$ vary across different methods. We compare APE with sequential encoding with and without RAG, and PCW, using \textsc{Llama-3-8B-Instruct}~\citep{llama3}, \textsc{Mistral-7B-Instruct-v0.3}~\citep{jiang2023mistral}, \textsc{Gemma-2-9b-it}~\citep{gemma2}, and \textsc{Llama-3.1-8B-Instruct} as base models. 



\textbf{Results.} In Table~\ref{tab:rag_longbench}, APE consistently improves performance across all models, achieving a 5.6\% average gain over sequential encoding without RAG. It also outperforms sequential RAG baselines by 3.3\% by retrieving more and longer contexts. The superior performance over PCW further showcases the effectiveness of our modifications in APE. Notably, APE surpasses the 128K-context variant of the \textsc{Llama-3.1-8B-Instruct} model by placing retrieved texts within the 8K context window, mitigating the ``lost in the middle" phenomenon.

\begin{table}[t]
\centering
\small
\caption{\textbf{\mbox{Comparison between APE and baselines on LongBench across different models using RAG.}} C denotes Contriever, and $M \times N$ indicates retrieval of the top-$N$
chunks, each containing $M$ words.}
\resizebox{\linewidth}{!}{\setlength{\tabcolsep}{0.5mm}{
\begin{tabular}{l|ccccccccc}
\toprule
Model & MuSiQue & Qasper & 2WikiMQA & DuRead & HotpotQA & NarratQA & MFQA\_zh & MFQA\_en & Avg. \\
\midrule
\textsc{LLaMA-3-8B-Instruct} & 20.70 & 41.05 & 30.02 & 9.55 & 45.90 & 20.98 & \textbf{58.54} & 45.04 & 33.97 \\
C200×20, Sequential & \textbf{27.93} & \textbf{42.71} & 38.35 & 12.65 & 49.60 & 22.78 & 57.82 & \textbf{48.94} & 37.60 \\
C4000×20, PCW & 18.82 & 42.59 & 40.99 & 21.57 & 47.09 & 23.29 & 54.40 & 45.05 & 36.73 \\
\rowcolor{cyan!10}C4000×20, APE & 26.19 & 42.32 & \textbf{44.43} & \textbf{23.13} & \textbf{49.71} & \textbf{30.71} & 55.03 & 45.41 & \textbf{39.62} \\
\midrule
\textsc{Mistral-7B-Instruct-v0.3} & 10.05 & 31.08 & 22.12 & 17.68 & 32.09 & 19.68 & 32.03 & 40.38 & 25.64 \\
C200×20, Sequential & 11.58 & 21.98 & 24.44 & 20.80 & 32.79 & 16.06 & 34.43 & 38.40 & 25.06 \\
C4000×20, PCW & 17.58 & 35.57 & 32.97 & 18.70 & 37.05 & 14.10 & 34.69 & 40.14 & 28.85 \\
\rowcolor{cyan!10}C4000×20, APE & \textbf{20.30} & \textbf{36.81} & \textbf{34.37} & \textbf{21.89} & \textbf{42.33} & \textbf{20.49} & \textbf{40.20} & \textbf{44.03} & \textbf{32.55} \\
\midrule
\textsc{Gemma-2-9b-it} & 22.57 & 39.99 & 48.06 & 27.40 & 47.49 & 23.11 & 50.81 & 45.35 & 38.10 \\
C200×10, Sequential & 30.69 & 42.86 & \textbf{53.55} & 28.04 & 52.05 & 24.45 & 50.25 & 48.34 & 41.28 \\
C2000×20, PCW & 26.27 & 46.69 & 47.59 & 23.43 & 48.95 & 27.11 & \textbf{56.69} & 49.81 & 40.82 \\
\rowcolor{cyan!10}C2000×20, APE & \textbf{33.38} & \textbf{47.72} & 49.49 & \textbf{28.43} & \textbf{56.62} & \textbf{30.41} & 56.52 & \textbf{50.84} & \textbf{44.18} \\
\midrule
\textsc{LLaMA-3.1-8B-Instruct} & 22.18 & \textbf{46.81} & 40.58 & \textbf{34.61} & 43.97 & 23.08 & 61.60 & 51.89 & 38.98 \\
\textcolor{darkgray!50}{128K, Sequential} & \textcolor{darkgray!50}{28.35} & \textcolor{darkgray!50}{47.20} & \textcolor{darkgray!50}{40.81} & \textcolor{darkgray!50}{33.34} & \textcolor{darkgray!50}{53.46} & \textcolor{darkgray!50}{30.57} & \textcolor{darkgray!50}{61.97} & \textcolor{darkgray!50}{53.25} & \textcolor{darkgray!50}{42.24} \\
C200×20, Sequential & \textbf{30.62} & 42.33 & 44.39 & 33.51 & 49.97 & 23.87 & 56.87 & \textbf{55.14} & 40.22 \\
C4000×20, PCW & 21.23 & 41.52 & 44.87 & 31.11 & 49.47 & 19.98 & 60.90 & 51.19 & 38.44 \\
\rowcolor{cyan!10}C4000×20, APE & 26.88 & 43.03 & \textbf{50.11} & 32.10 & \textbf{55.41} & \textbf{30.50} & \textbf{62.02} & 52.51 & \textbf{42.86} \\
\bottomrule
\end{tabular}}}
\label{tab:rag_longbench}
\end{table}

\subsection{In-context Learning}
\label{icl}

\textbf{Setup.} We evaluate APE on three ICL tasks using the LM Evaluation Harness~\citep{eval-harness} codebase: GSM8K (8-shot)~\citep{cobbe2021training}, TriviaQA (5-shot)~\citep{joshi2017triviaqa}, and MMLU (5-shot)~\citep{hendrycks2020measuring}. Experiments are conducted using the same base models as in our LongBench evaluations. We compare parallel encoding (PCW) to show the improvement of APE. Sequential encoding with varying numbers of shots (i.e., 1-shot, half-shots, and full-shots) is also employed to measure the gap from the ideal scenarios.

\begin{figure}[t]
    \centering
    \begin{subfigure}[b]{0.49\textwidth}
        \centering
        \includegraphics[width=\textwidth]{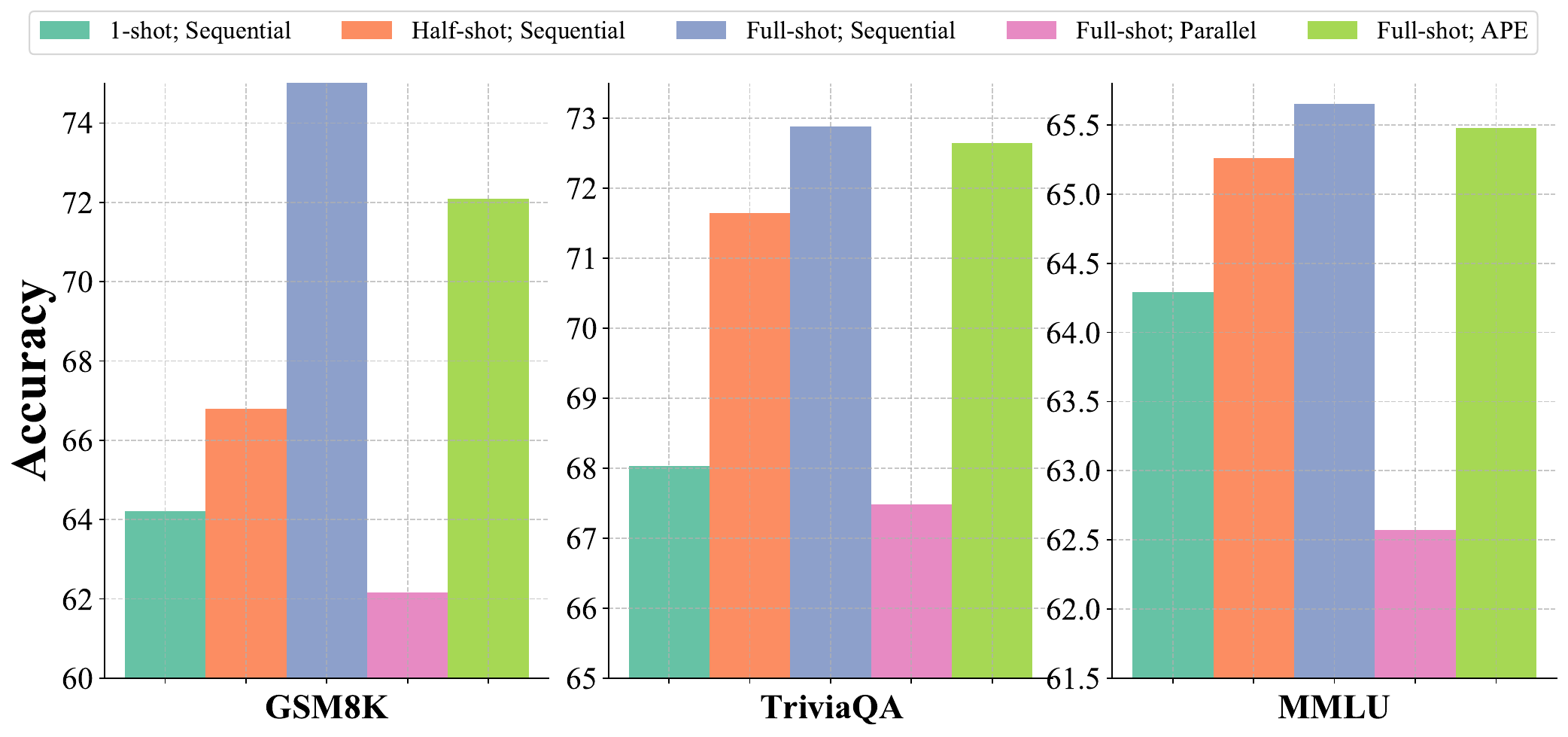}
        \caption{\textsc{Llama-3-8B-Instruct}}
        \label{fig:subfig1}
    \end{subfigure}
    \hfill
    \begin{subfigure}[b]{0.49\textwidth}
        \centering
        \includegraphics[width=\textwidth]{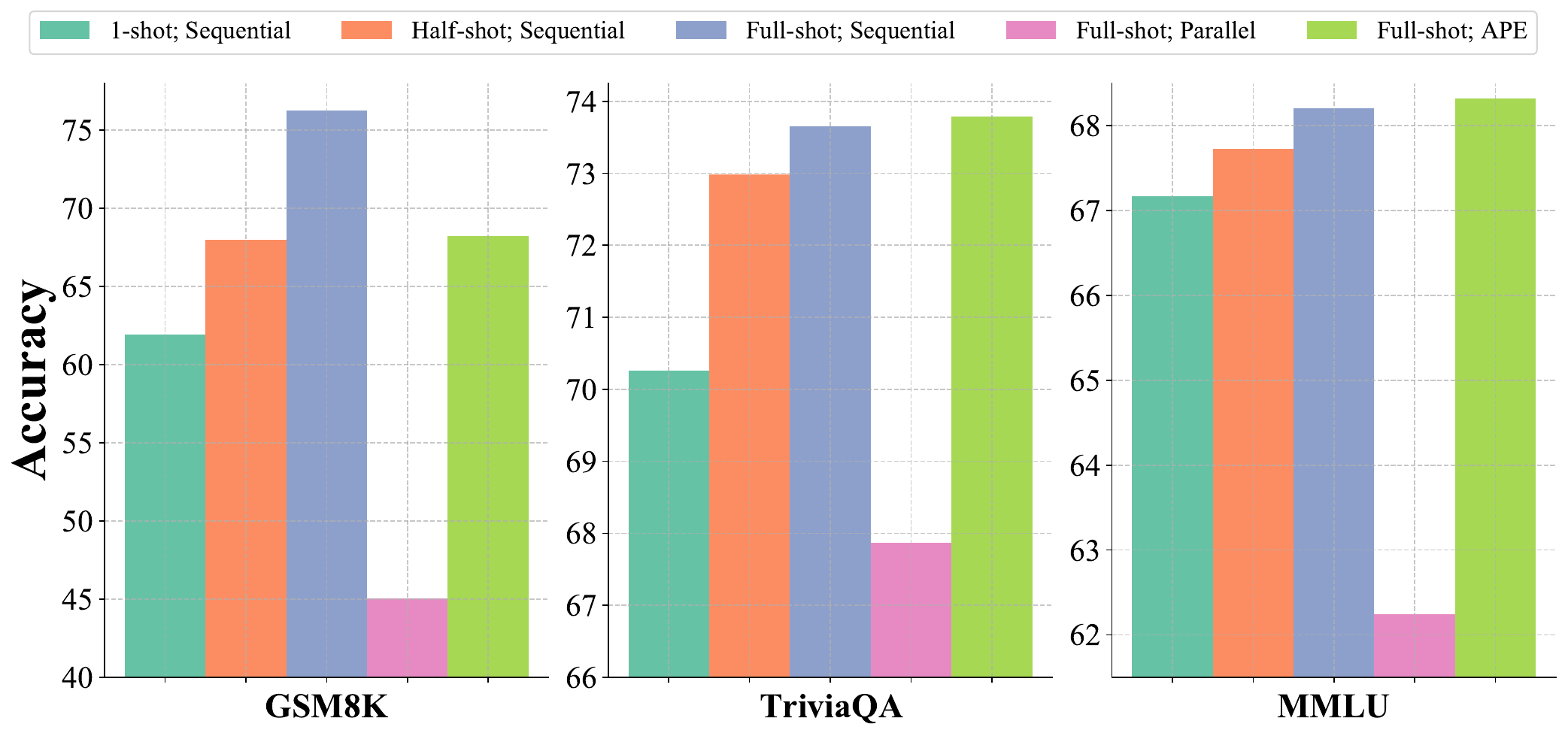}
        \caption{\textsc{Llama-3.1-8B-Instruct}}
        \label{fig:subfig2}
    \end{subfigure}
    \vskip\baselineskip
    \begin{subfigure}[b]{0.49\textwidth}
        \centering
        \includegraphics[width=\textwidth]{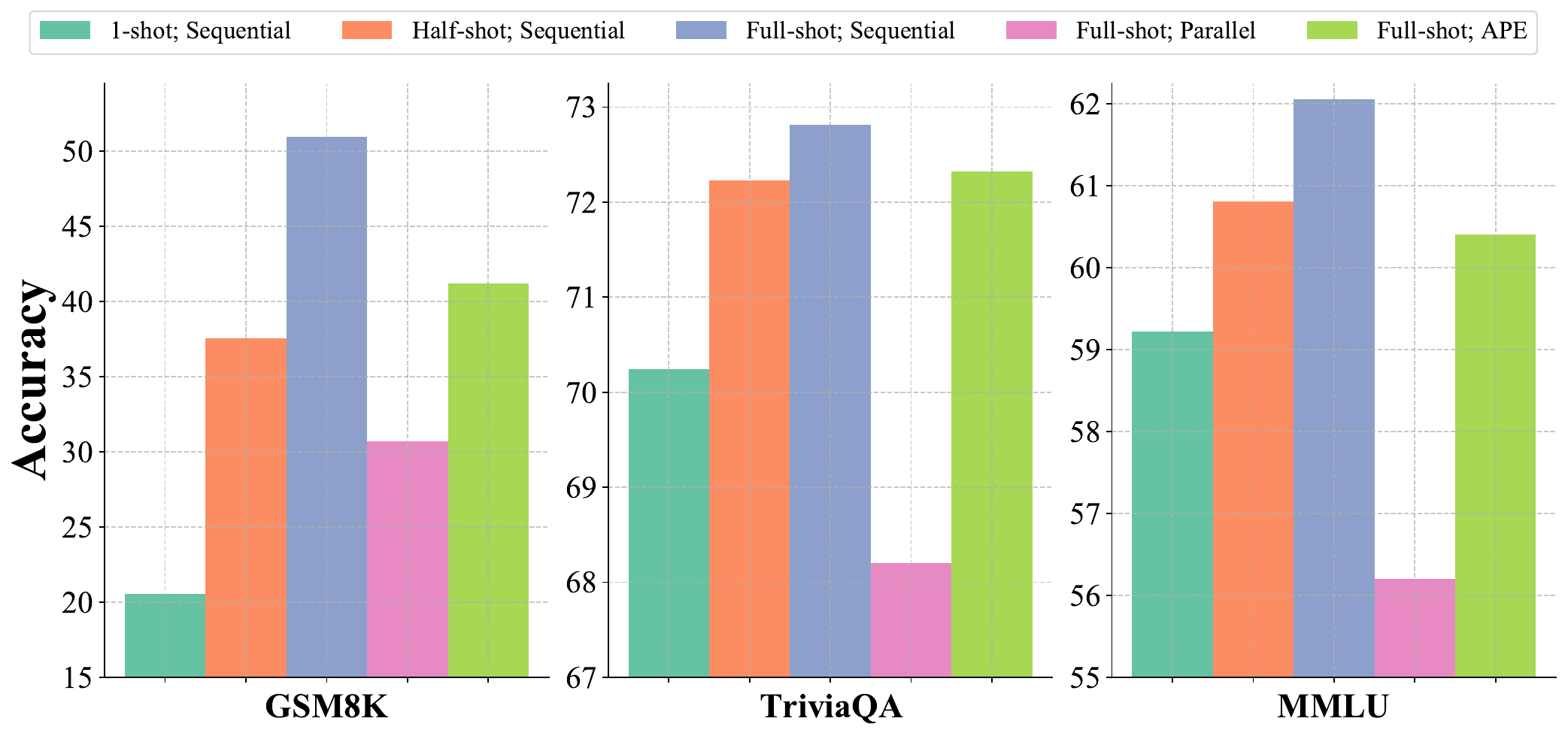}
        \caption{\textsc{Mistral-7B-Instruct-v0.3}}
        \label{fig:subfig3}
    \end{subfigure}
    \hfill
    \begin{subfigure}[b]{0.49\textwidth}
        \centering
        \includegraphics[width=\textwidth]{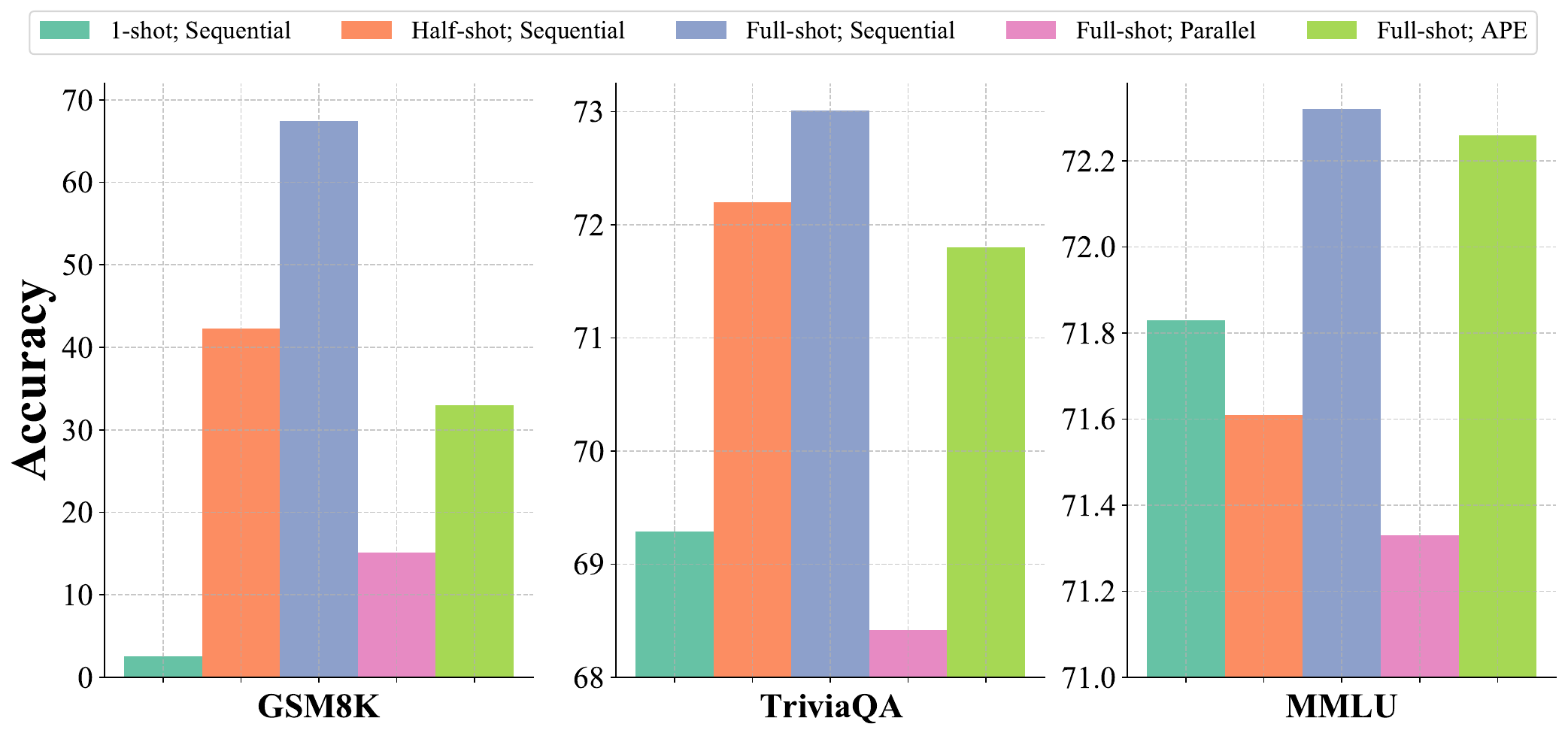}
        \caption{\textsc{Gemma-2-9b-it}}
        \label{fig:subfig4}
    \end{subfigure}
    \caption{\textbf{Performance comparison of APE, parallel encoding, and sequential encoding on ICL tasks.}}
    \label{fig:icl}
\end{figure}

\textbf{Results.} In Figure~\ref{fig:icl}, APE surpasses parallel encoding with average improvements of 15.4\% on GSM8K, 4.7\% on TriviaQA, and 3.5\% on MMLU. When compared with the 1-shot sequential baseline with similar context length, our method consistently yields superior results. Moreover, APE performs better than half-shot sequential encoding in 8/12 settings and preserves 93\% accuracy compared to full-shot sequential encoding. Additionally, the \textsc{Llama} family exhibits enhanced compatibility with parallel encoding, potentially due to the stronger directional alignment of initial tokens from different contexts (see Figure~\ref{fig:observation2}a). Across different tasks, the performance gap between APE and full-shot sequential encoding is the largest on GSM8K. This finding suggests that while APE keeps most capabilities, its effectiveness may decrease as task complexity increases.

\subsection{Many-shot Context-Augmented Generation}
\label{sec:loft}

\begin{table}[ht]  
\caption{\textbf{Comparison between APE and sequential encoding in various many-shot RAG and ICL tasks.}}
\resizebox{\linewidth}{!}{\setlength{\tabcolsep}{2mm}{
\begin{tabular}{l|cccc|cccc}
\toprule
 & \multicolumn{4}{c|}{Retrieval-augmented Generation} & \multicolumn{4}{c}{In-context Learning} \\  
\cmidrule{0-8}
Method & ArguAna & FEVER & NQ & SciFact & Date & Salient & Tracking7 & Web \\ 
\midrule
Sequential, Zero-shot & 11.15 & 7.78 & 17.78 & 7.74 & 20.00 & 8.89 & 1.12 & 8.89 \\  
Sequential, Few-shot & 11.20 & 9.78 & 17.81 & 9.49 & 36.64 & 38.89 & 6.67 & 38.89 \\
Sequential, Half-shot & 15.34 & 13.12 & 19.64 & 16.12 & 45.55 & 42.22 & 8.89 & 55.56 \\
Sequential, Full-shot & 12.84 & 14.19 & \textbf{24.54} & \textbf{16.88} & \textbf{46.67} & \textbf{46.67} & \textbf{8.89} & \textbf{58.89} \\
\midrule
\rowcolor{cyan!10}APE, Full-shot & \textbf{16.32} & \textbf{14.70} & 21.91 & 15.72 & 43.33 & 45.55 & \textbf{8.89} & \textbf{58.89} \\
\bottomrule
\end{tabular}}}
\label{tab:loft}
\end{table}

\textbf{Setup.} We evaluate the scalability of APE on four RAG and ICL tasks from the LOFT benchmark~\citep{lee2024can}, each involving hundreds of additional texts. We employ \textsc{Llama-3.1-8B-Instruct} as our base model to compare APE with sequential encoding, both applied to the same many-shot inputs. The total context lengths for the RAG and ICL tasks are 128K and 32K, respectively. We also include the zero-shot, few-shot ($\leq$ 5), and half-shot sequential encoding baselines. For metrics, F1 score and EM are used in RAG and ICL tasks.

\textbf{Results.} In Table~\ref{tab:loft}, APE achieves performance comparable to sequential encoding when processing the same many-shot long-context inputs, showing its ability to encode hundreds of texts in parallel efficiently. Notably, it outperforms sequential encoding on ArguAna and FEVER for RAG tasks. While APE is expected to reduce performance, it recovers this drop by positioning all texts close to the query, mitigating the ``lost in the middle" problem in long-context LLMs. For ICL tasks, APE can learn from examples as effective as sequential encoding.

\subsection{Efficiency Evaluation}
\label{sec:efficiency}

\begin{figure}[ht]
    \centering
    \begin{subfigure}[b]{0.245\textwidth}
        \centering
        \includegraphics[width=\textwidth]{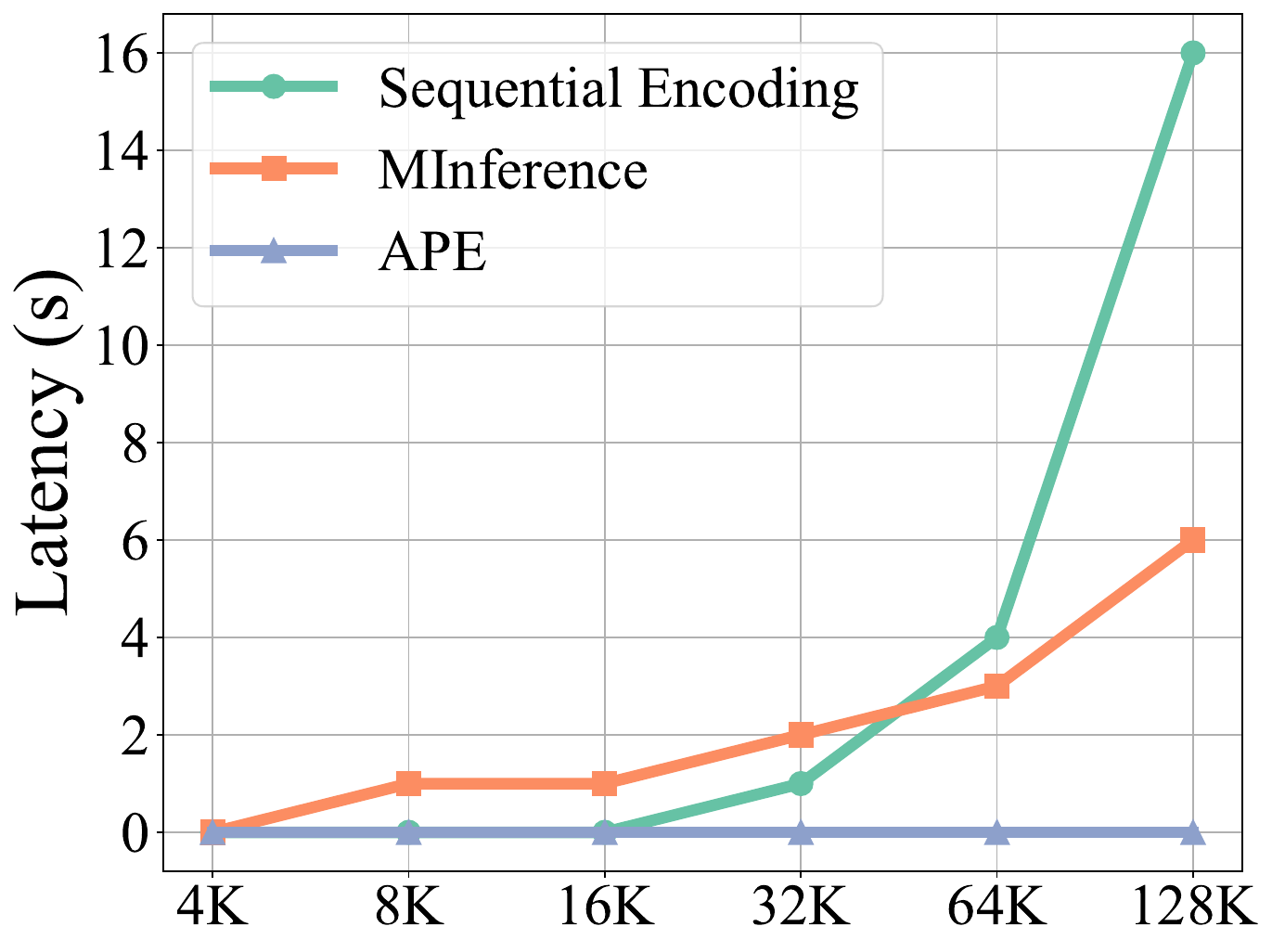}
        \caption{Prefill Time (\color{gray}{bsz=1})}
    \end{subfigure}
    \hfill
    \begin{subfigure}[b]{0.23\textwidth}
        \centering
        \includegraphics[width=\textwidth]{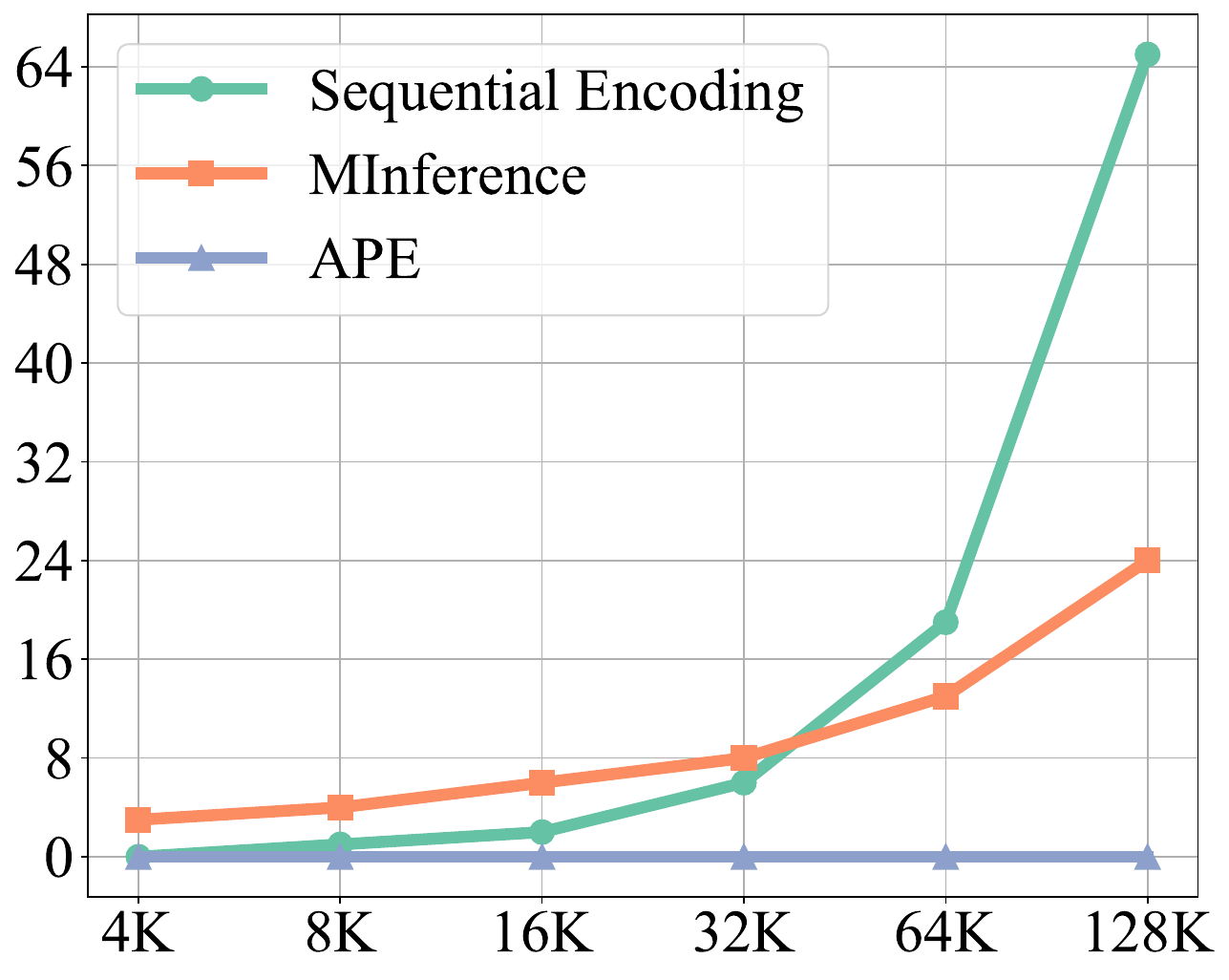}
        \caption{Prefill Time (\color{gray}{bsz=4})}
    \end{subfigure}
    \hfill
    \begin{subfigure}[b]{0.23\textwidth}
        \centering
        \includegraphics[width=\textwidth]{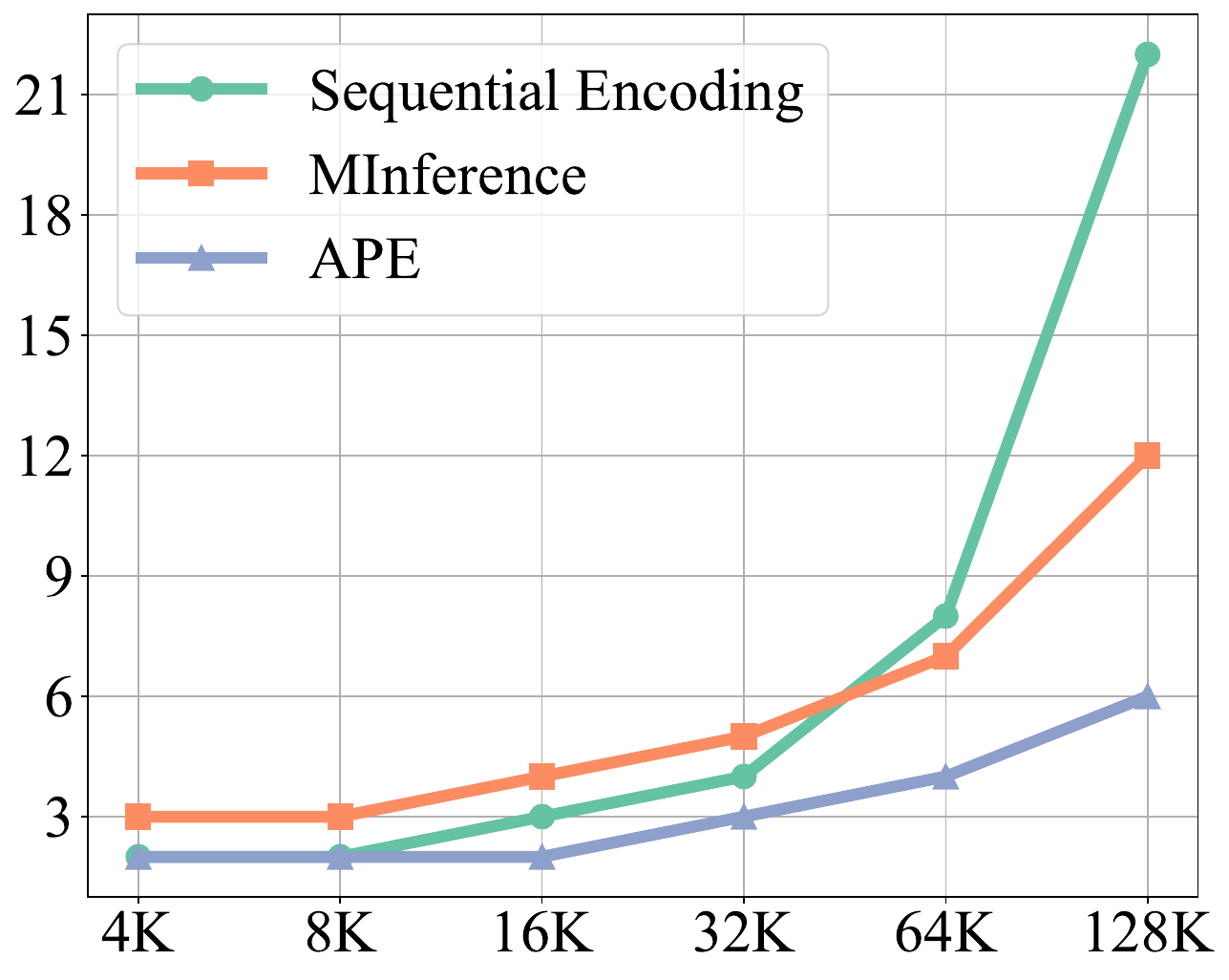}
        \caption{Total Time (\color{gray}{bsz=1})}
    \end{subfigure}
    \hfill
    \begin{subfigure}[b]{0.23\textwidth}
        \centering
        \includegraphics[width=\textwidth]{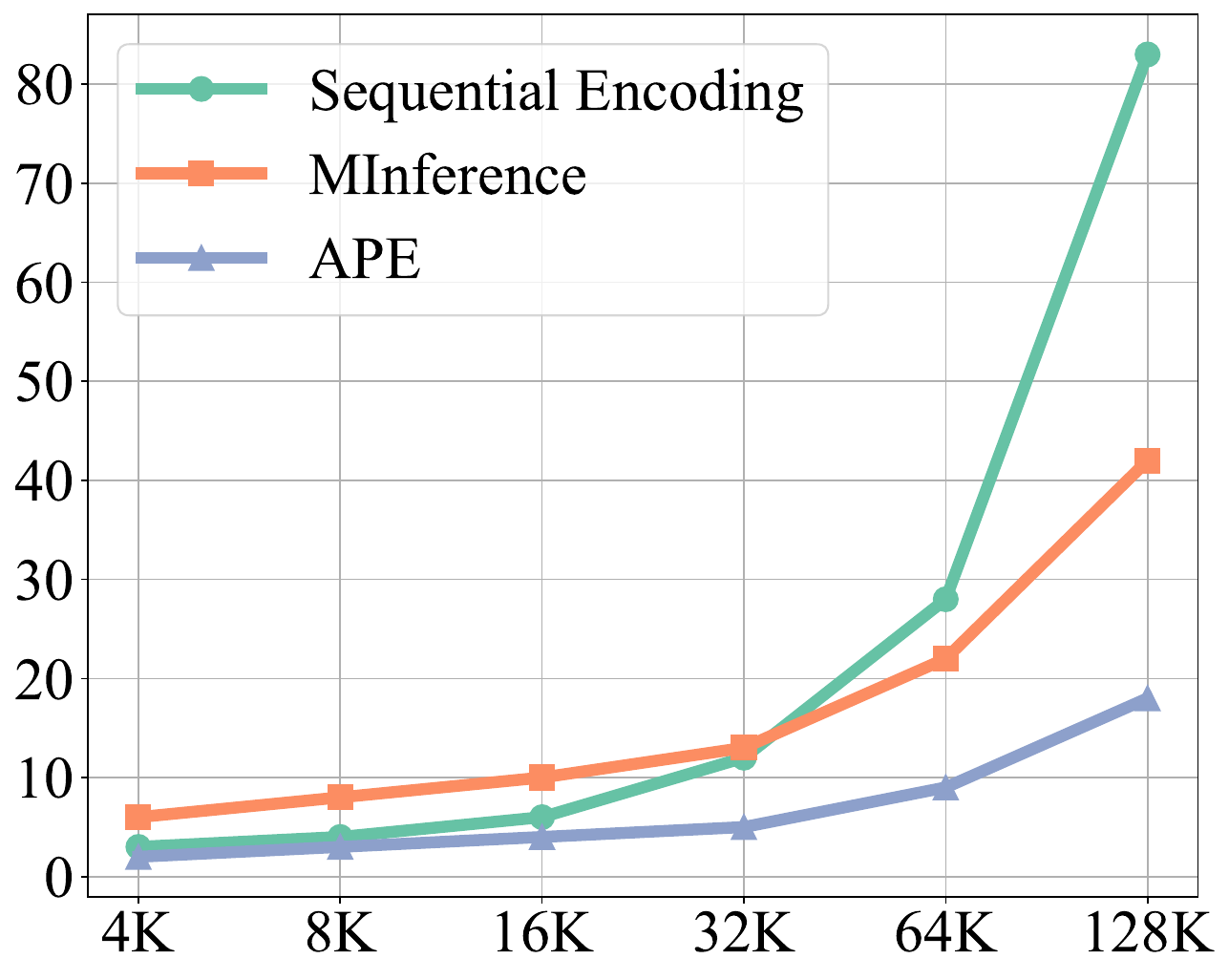}
        \caption{Total Time (\color{gray}{bsz=4})}
    \end{subfigure}
    \caption{\textbf{Latency on H100 GPU: prefill and total inference time (s).} The gray text in brackets is batch size.} 
    \label{fig:latency}
\end{figure}

\textbf{Setup.} We measure the latency for sequential encoding, MInference~\citep{jiang2024minference}, and APE usingLlama-3.1-8B-Instruct~\citep{llama3} on an H100 GPU with batch sizes of 1 and 4. The query and generation lengths are fixed at 256 tokens, while the context lengths range from 2K to 128K tokens. We employ VLLM~\citep{kwon2023efficient} as our inference engine and measure both prefilling time and total inference time.

\textbf{Results.} Comparing to sequential encoding and MInference,  APE can accelerate inference up to 4.5$\times$ and 2.2$\times$ respectively for long-context scenarios in Figure~\ref{fig:latency}. For 128K-token contexts, APE reduces prefilling time by 28$\times$ compared to MInference. The prefilling cost of APE exhibits linear scaling and consumes less than 10\% of inference time, whereas baselines require over 50\% as context length increases. APE also shows superior versatility, while MInference slows inference with additional overhead for short contexts and large batches.
\section{Analysis} 

This section presents analyses to answer the following research questions: \textbf{RQ1}: Can APE improve performance for real-world RAG applications? \textbf{RQ2}: How does each component in APE contribute to the performance? \textbf{RQ3}: Can APE extend LLM context window size in long-context scenarios without RAG?

\subsection{Can APE improve performance for real-world RAG applications?}

In Table~\ref{tab:crag}, we evaluate APE in real-world RAG scenarios using the CRAG benchmark \citep{yang2024crag}. Task 1 augments the model with five webpages, while Task 2 provides an additional knowledge graph as another retrieval source. In our experiments, the sequential encoding baseline is limited to retrieving 4K tokens, whereas APE can process 20 parallel segments of 4K tokens each. By incorporating significantly more external texts, APE consistently outperforms sequential encoding with limited context sizes while reducing latency. Moreover, the improvement in Task 2 shows the effectiveness of APE in merging text from multiple sources.

\begin{table}[h]
\centering
\small
\caption{\textbf{Performance and latency comparison using the Llama-3-8B-Instruct model on CRAG benchmark.}}
\vspace{-0.8em}
\resizebox{\linewidth}{!}{\setlength{\tabcolsep}{1mm}{
\begin{tabular}{llccccc}
\toprule
Task & Model & Latency (ms) & Accuracy (\%)  & Hallucination & Missing & Score$_a$ \\
\midrule
LLM only & \textsc{Llama-3-8B-Instruct} & 682 & 22.14 & 48.97 & 28.90 & -26.83 \\
 \midrule
\multirow{2}{*}{Task 1} & \textsc{Llama-3-8B-Instruct} & 1140 & 23.28 & 29.49 & 47.22 & -6.21 \\
 & \multicolumn{1}{>{\columncolor{cyan!10}}l}{+APE} & \multicolumn{1}{>{\columncolor{cyan!10}}c}{1054} & \multicolumn{1}{>{\columncolor{cyan!10}}c}{\textbf{25.53}} & \multicolumn{1}{>{\columncolor{cyan!10}}c}{\textbf{21.30}} & \multicolumn{1}{>{\columncolor{cyan!10}}c}{\textbf{37.93}} & \multicolumn{1}{>{\columncolor{cyan!10}}c}{\textbf{-0.41}} \\
  \midrule
\multirow{2}{*}{Task 2} & \textsc{Llama-3-8B-Instruct} & 1830 & 24.46 & 28.38 & 47.15 & -3.92 \\
 & \multicolumn{1}{>{\columncolor{cyan!10}}l}{+APE} & \multicolumn{1}{>{\columncolor{cyan!10}}c}{1672} & \multicolumn{1}{>{\columncolor{cyan!10}}c}{\textbf{27.04}} & \multicolumn{1}{>{\columncolor{cyan!10}}c}{\textbf{18.74}} & \multicolumn{1}{>{\columncolor{cyan!10}}c}{\textbf{37.32}} & \multicolumn{1}{>{\columncolor{cyan!10}}c}{\textbf{2.16}} \\
 \bottomrule
\end{tabular}}}
\label{tab:crag}
\end{table}

\subsection{How does each component in APE contribute to the performance?}

\begin{wraptable}{r}{0.4\textwidth}
\centering
\vspace{-1.4em}
\caption{\textbf{Ablation study of APE components on ICL tasks.} $P$: shared prefix, $T$: attention temperature, $S$: scaling factor.}
\vspace{-0.4em}
\resizebox{\linewidth}{!}{\setlength{\tabcolsep}{1mm}{
\begin{tabular}{ccc|ccc}
\toprule
$P$ & $T$ &  $S$ & GSM8K & TriviaQA & MMLU \\  \midrule
  &  & & 38.25\% & 67.99\% & 63.09\%  \\
\checkmark &  &  & 50.42\% & 70.76\% & 63.70\%  \\
\checkmark & \checkmark &  & 51.15\% & 71.03\%  & 64.49\%   \\\midrule
 \checkmark & \checkmark & \checkmark & \textbf{53.62\%} & \textbf{72.64\%} & \textbf{66.62\%}   \\
\bottomrule
\end{tabular}}}
\vspace{-2em}
\label{fig:ablation}
\end{wraptable}

In Table~\ref{fig:ablation}, we conduct an ablation study to examine each component in APE, including the shared prefix ($P$), attention temperature ($T$), and scaling factor ($S$). We present results averaged across the four base models evaluated in Figure~\ref{fig:icl}. Our findings indicate that incorporating each of these components can improve performance for all tasks, with average improvements of 5.19\%, 0.59\%, and 2.07\%, respectively. Among them, adding a shared prefix leads to the largest improvement, while adjusting the attention temperature yields minimal accuracy gains without the complementary effect of the scaling factor.

\subsection{Can APE extend context lengths in long-context scenarios without RAG?}

Table~\ref{tab:single} evaluates the effectiveness of APE in handling a single long-context input using the \textsc{Llama-3-8B-Instruct} model on the LongBench dataset~\citep{bai2023longbench}. To accommodate the long context within our APE, we split it into multiple segments of less than 7,500 tokens. Additionally, we append the last 500 tokens to the query for two code completion tasks. Our results indicate that APE enhances performance across 10/11 tasks, yielding an average improvement of 6.6\% compared to the sequential encoding baseline with limited context window size. More baseline results of long-context LLM approaches are provided in Appendix~\ref{sec:app:lclm}.

\begin{table}[h]
\centering
\caption{\textbf{Performance comparison across different long-context tasks on LongBench~\citep{bai2023longbench}.}}
\vspace{-0.7em}
\resizebox{\linewidth}{!}{\setlength{\tabcolsep}{1mm}{
\begin{tabular}{l|cccccc}
\toprule
Method & NarratQA & Qasper & MultiFQA & GovReport & QMSum & LCC \\ \midrule
\textsc{LLaMA-3-8B-Instruct} & 19.32 & 32.83 & 43.38 & 27.89 & 22.40 & 53.22 \\
\rowcolor{cyan!10}+APE & \textbf{26.87} & \textbf{39.14} & \textbf{59.12} & \textbf{29.10} & \textbf{23.08} & \textbf{66.09} \\
\midrule\midrule
Method & RepoBench-P & HotpotQA & 2WikiMQA & MuSiQue & MultiNews & Average \\ \midrule
\textsc{LLaMA-3-8B-Instruct} & 38.15 & 44.24 & 21.01 & 20.47 & \textbf{23.63} & 31.50 \\
\rowcolor{cyan!10}+APE & \textbf{49.43} & \textbf{50.11} & \textbf{28.06} & \textbf{25.79} & 22.40 & \textbf{38.11} \\
\bottomrule
\end{tabular}}}
\label{tab:single}
\end{table}

\section{Conclusion}
\label{Conclusion}
This work explores the potential of parallel encoding in CAG scenarios, which can pre-cache KV states for fast inference and re-use positions for long context but lead to worse performance. To address this, we propose APE, a training-free method to enable accurate, fast, and long CAG systems. APE achieves this by aligning the attention weight distribution of parallel encoding with sequential encoding via three steps: shared prefix, adaptive temperature, and scaling factor. Empirically, we show that APE improves accuracy and efficiency in various RAG and ICL tasks while successfully scaling to process hundreds of chunks in parallel for both settings.

\section{Limitations}

While APE shows the effectiveness and efficiency of parallel encoding with only inference-time modification in the attention distribution, it remains sensitive to hyperparameter selection, particularly the attention temperature $T$ and scaling factor $S$. 
In real-world applications, where contexts vary in length, quantity, and content, aligning the distribution between sequential and parallel encoding automatically presents a significant challenge. 

\section{Acknowledgement}

This work is supported in part by NSF award CNS-2211882 and a gift from Qualcomm. We thank the authors of ChatQA~\citep{liu2024chatqa}, Longbench~\citep{bai2023longbench}, CRAG~\citep{yang2024crag}, LM Evaluation Harness~\citep{eval-harness}, VLLM~\citep{kwon2023efficient}, and MInference~\citep{jiang2024minference} for their useful codebase, benchmark, and models, and Yixin Dong, Hanshi Sun, Zhuoming Chen for their helpful discussions.

\clearpage
\newpage
\bibliographystyle{assets/plainnat}
\bibliography{paper}

\clearpage
\newpage
\beginappendix
\section{Detailed Experimental Setups for Section~\ref{obs2}}
\label{app:obs1}

\textbf{RAG.} We select four tasks that require processing multiple input documents from the LongBench dataset~\citep{bai2023longbench}, including HotpotQA~\citep{yang2018hotpotqa}, 2WikiMultihopQA~\citep{ho2020constructing}, MuSiQue~\citep{trivedi2022musique}, and MultiNews~\citep{fabbri2019multi}. The F1 score is used as the evaluation metric for the three QA tasks, while Rouge-L is used for the summarization task. Both parallel encoding and CEPED process each document independently using $\Theta_{\text{Enc}}$. For documents that exceed the length limitation of $\Theta_{\text{Enc}}$, we split them into multiple chunks for encoding. In sequential encoding, we will truncate lengthy inputs from the middle.

\textbf{ICL.} We select three few-shot learning tasks from LM Evaluation Harness~\citep{eval-harness} to evaluate the ICL ability of different encoding methods, involving GSM8K~\citep{gsm8k}, TriviaQA~\cite{joshi2017triviaqa}, and MMLU~\citep{mmlu}. In parallel encoding and CEPED, we will encode each example separately and input all the resulting KV states to $\Theta_{\text{Dec}}$. For sequential encoding, we use variants with different numbers of shots to further measure the effectiveness of other methods, including 0-shot, 1-shot, half-shot, and full-shot.

\section{More Visualization Results for Section 3.2}
\label{app:obs2}

\subsection{Similarity between Tokens from Different Samples in Each Position for Key States.}

\begin{figure}[ht]
    \centering
    \includegraphics[width=\textwidth]{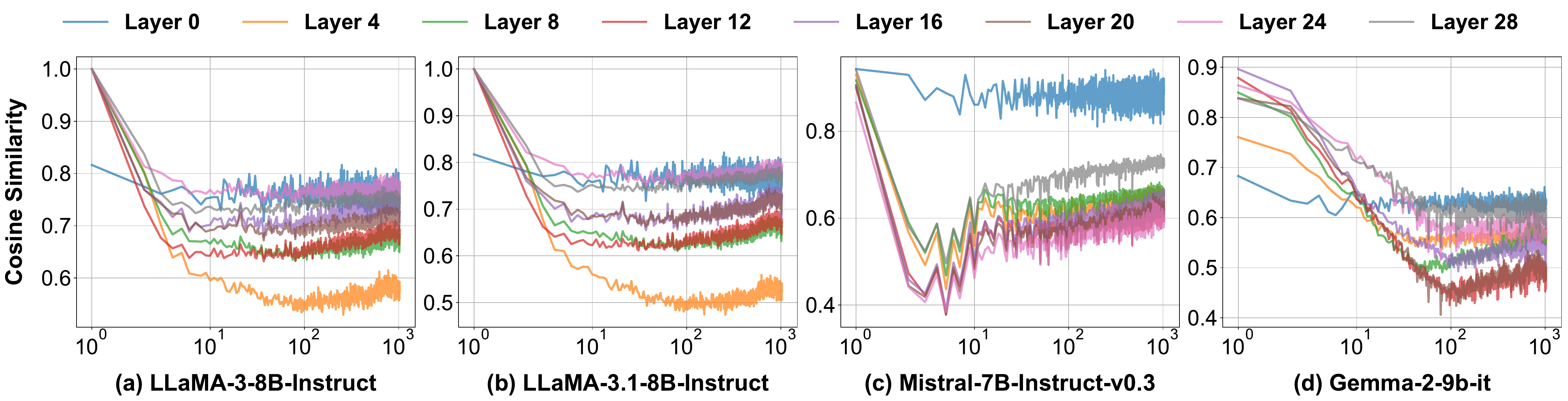}
    \caption{For all base models, key states from distinct inital tokens exhibit a large cosine similarity than the following positions, where the LLaMA family even approaches 1. The X-axis shows positions of key states on a logarithmic scale.} 
    \label{fig:app:sim:k}
\end{figure}

In Figure~\ref{fig:app:sim:k}, we showcase that key states in different layers maintain consistently high cosine similarity values for various initial tokens, with only the first layer exhibiting slightly lower similarities. Our analysis reveals that \textsc{LLaMA-3-8B-Instruct} and \textsc{LLaMA-3.1-8B-Instruct} exhibit almost the same direction (approximately 1.0) for different tokens beyond the first layer, while \textsc{Mistral-7B-Instruct-v0.3} and \textsc{Gemma-2-9b-it} show substantial but lower similarities ranging from 0.8 to 0.9. These findings indicate inherent alignments across contexts while highlighting the potential for further improvements through the shared prefix in Section~\ref{sec:shared_prefix}.

\subsection{Similarity between Tokens from Different Samples in Each Position for Value States.}

\begin{figure}[ht]
    \centering
    \includegraphics[width=\textwidth]{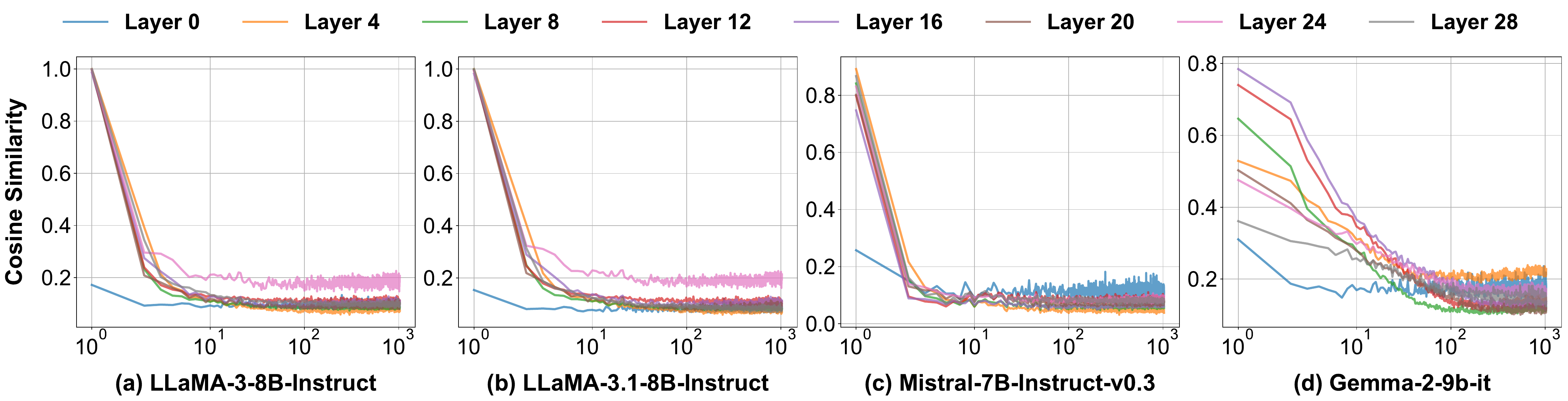}
    \caption{Among four models, value states from distinct inital tokens exhibit a large cosine similarity than the following positions, except the first layer and \textsc{Gemma-2-9b-it}. The X-axis shows positions of value states on a logarithmic scale.} 
    \label{fig:app:sim:v}
\end{figure}

Similarly, Figure~\ref{fig:app:sim:v} shows that value states maintain high cosine similarity across different layers for various initial tokens. There are two notable exceptions: the first layer and the \textsc{Gemma-2-9b-it} model. This distinctive pattern in \textsc{Gemma-2-9b-it} aligns with the model's requirement for a system prompt to function correctly.

\subsection{Similarity between the Initial Token and Following Tokens for Key States.}

\begin{figure}[ht]
    \centering
    \includegraphics[width=\textwidth]{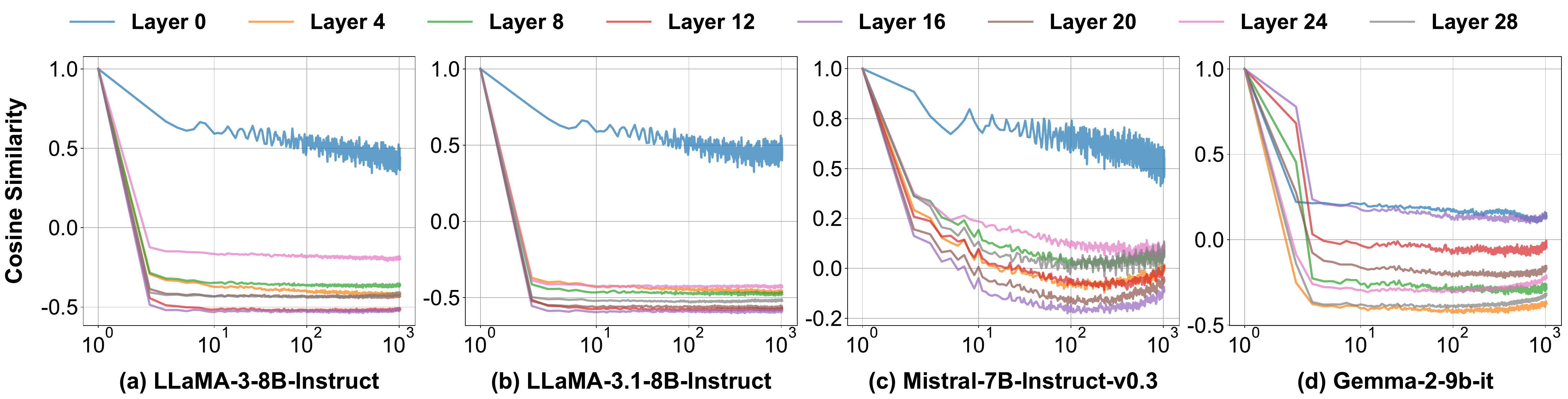}
    \caption{For all base models, the similarity between the initial key state and subsequent key states stabilizes as the position increases. The X-axis shows positions of key states on a logarithmic scale.} 
    \label{fig:app:sim:kk}
\end{figure}

Figure~\ref{fig:app:sim:kk} illustrates how the cosine similarity between the initial and subsequent key states stabilizes as position increases. This similarity converges to a near-constant value for all base models after 10 tokens. 

\subsection{Similarity between the Initial Token and Following Tokens for Value States.}

\begin{figure}[ht]
    \centering
    \includegraphics[width=\textwidth]{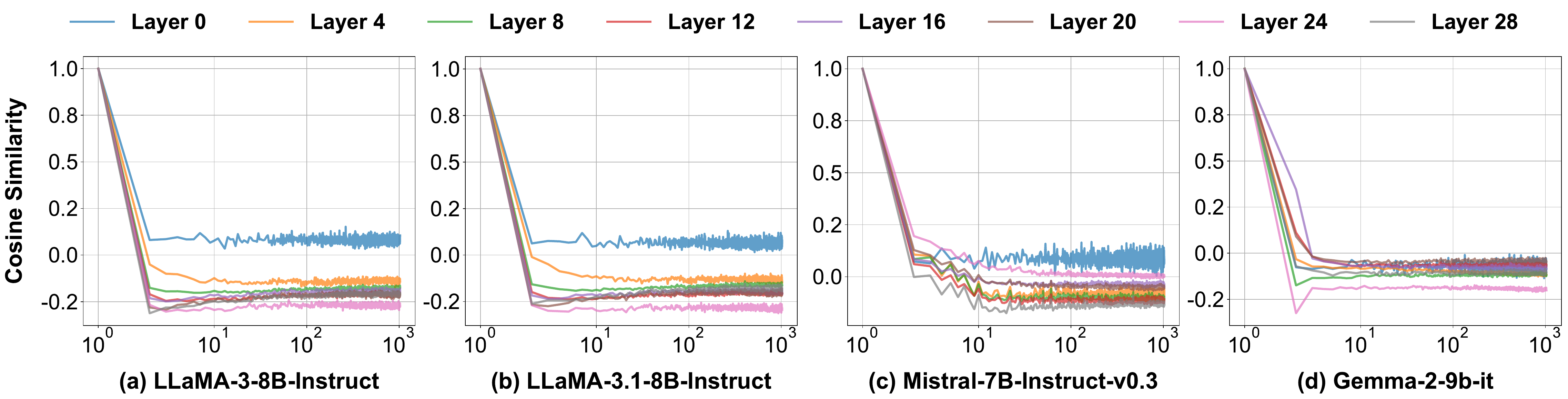}
    \caption{For all base models, the similarity between the initial value state and subsequent value states stabilizes as the position increases. The X-axis shows positions of value states on a logarithmic scale.} 
    \label{fig:app:sim:vv}
\end{figure}

Similar to key states, the value states exhibit a stable similarity between the initial token and subsequent tokens in Figure~\ref{fig:app:sim:vv}, with all models convergent to a nearly constant value after approximately 10 tokens.

\subsection{Similarity between the Query State and Past Key States.}

\begin{figure}[ht]
    \centering
    \includegraphics[width=\textwidth]{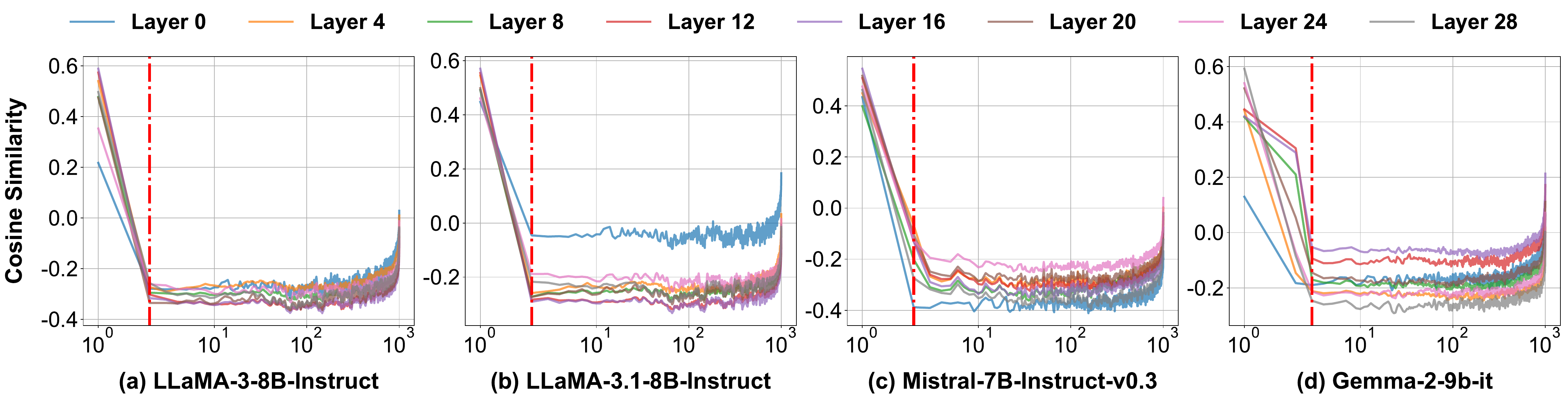}
    \caption{For all base models, the cosine similarity between the query state and past key states stabilizes for most positions, except for the initial and recent key states. The X-axis shows positions of key states on a logarithmic scale.} 
    \label{fig:app:sim:qk}
\end{figure}

In Figure~\ref{fig:app:sim:qk}, the query states across all layers, and base models exhibit higher cosine similarity with the initial tokens. Additionally, neighboring positions tend to receive higher cosine similarity.

\subsection{Magnitude of Key States from Different Positions.}

\begin{figure}[ht]
    \centering
    \includegraphics[width=\textwidth]{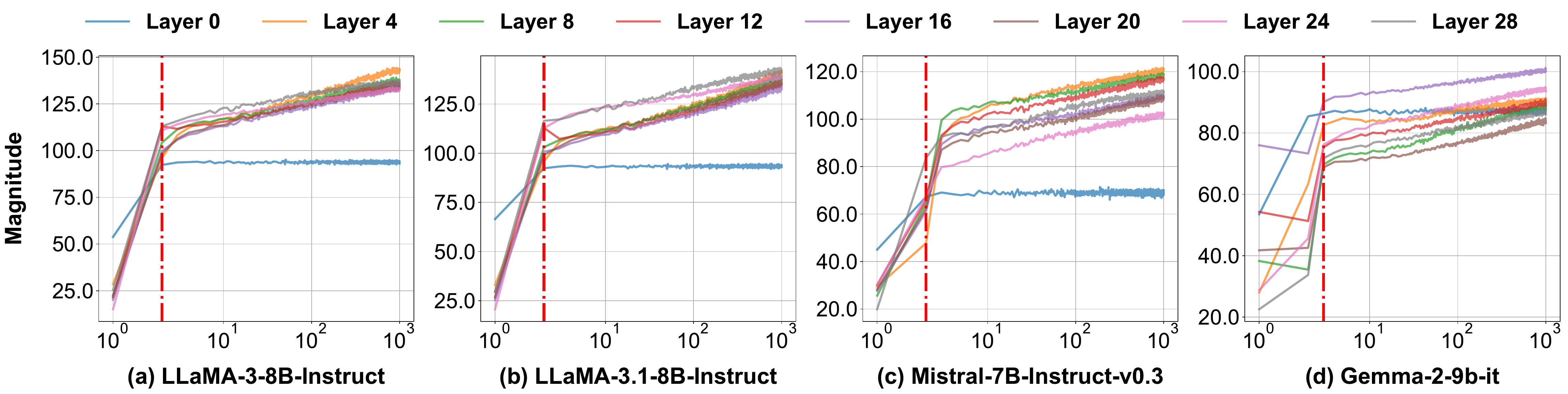}
    \caption{For all models, key states show a slowly upward trend in magnitude as position increases. A red dashed line marks the anomalous region for the first few tokens. The X-axis shows positions of key states on a logarithmic scale.} 
    \label{fig:app:norm:k}
\end{figure}

Figure~\ref{fig:app:norm:k} illustrates that the magnitude of key states gradually increases with position, except for the first few tokens, which exhibit significantly smaller magnitudes.

\subsection{Magnitude of Value States from Different Positions.}

\begin{figure}[ht]
    \centering
    \includegraphics[width=\textwidth]{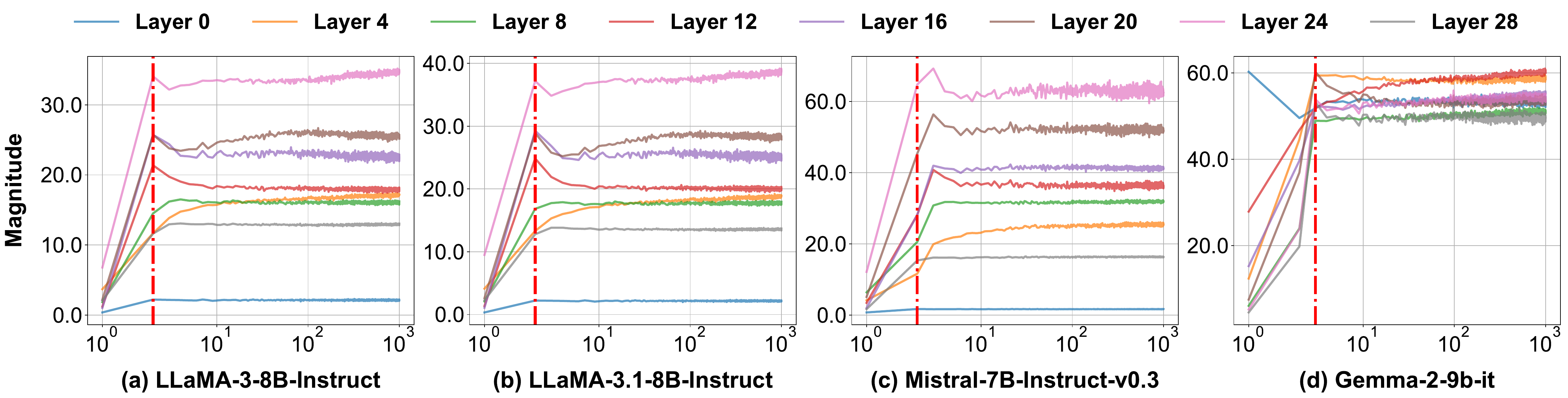}
    \caption{For all models, the magnitude of value states remains consistent for most positions, except for the first few positions highlighted by a red dashed line. The X-axis represents the positions of value states on a logarithmic scale.} 
    \label{fig:app:norm:v}
\end{figure}

In Figure~\ref{fig:app:norm:v}, the value states across all positions exhibit a similar magnitude, except for the first few positions, which show a noticeable deviation. We indicate this region with a red dashed line.

\subsection{Dot Product between the Query State and Past Key States.}

\begin{figure}[ht]
    \centering
    \includegraphics[width=\textwidth]{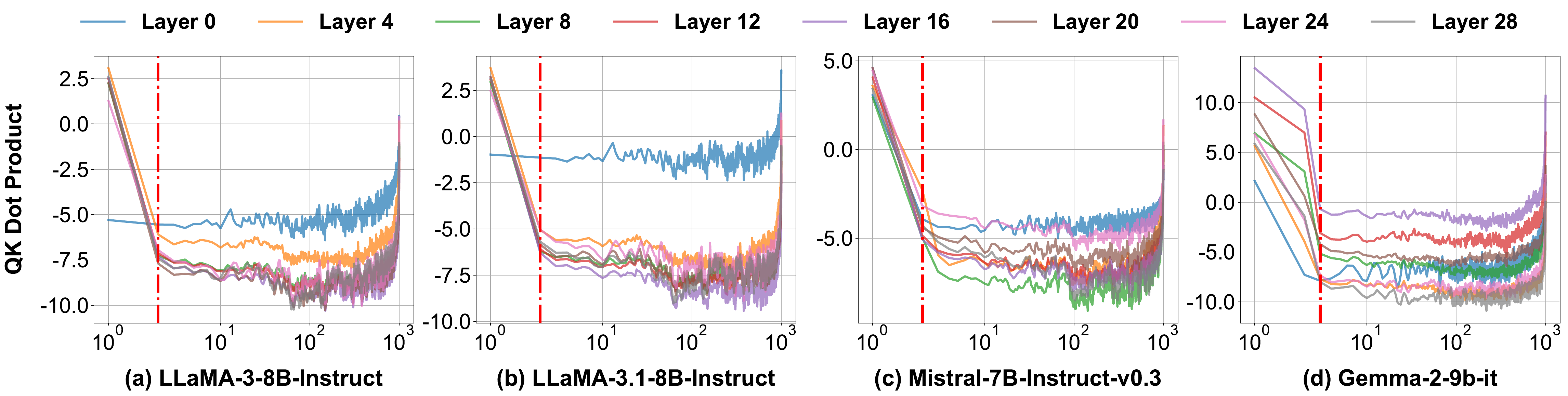}
    \caption{For all base models, the dot product values between the query state and past key states stabilizes for most positions, except for the initial and recent key states. The X-axis shows positions of key states on a logarithmic scale.} 
    \label{fig:app:sim:qk_score}
\end{figure}

In Figure~\ref{fig:app:sim:qk_score}, the query states across all layers, and base models exhibit larger dot product values with the initial tokens. Additionally, neighboring positions also tend to receive larger values.

\section{Formal Derivation of APE}
\label{app:algorithm}

\subsection{Hierarchical Formula for Softmax Attention.}

Here, we begin with the standard $\mathrm{Softmax}$ attention, where $Q$, $K$, and $V$ are the query, key, and value states from the input, respectively. To distinguish different sources, we use the subscript ${C_i}$ for elements originating from the context, while those without a subscript correspond to user queries or generated texts.
\begin{align}
O &= \mathrm{Softmax}\left(\frac{Q[K_{C_1}^\top, \ldots, K_{C_N}^\top, K^\top]}{\sqrt{d}}\right) \times [V_{C_1}, \ldots, V_{C_N}, V] \\
&= \frac{ [A_{C_1}, \ldots, A_{C_N}, A]}{\sum_{i=1}^{N}\sum_{j=1}^{l_{C_i}}a_{C_{i}, j}+ \sum_{j=1}^{l}a_{j}}\times [V_{C_1}, \ldots, V_{C_N}, V], \\
&\text{where }K_{C_i} = [k_{C_i, 1}, ..., k_{C_i, l_{C_i}}], V_{C_i} = [v_{C_i, 1}, ..., v_{C_i, l_{C_i}}], A_{C_i} = [\exp \frac{Qk_{C_i, 1}^\top}{\sqrt{d}}, \ldots, \exp \frac{Qk_{C_i, l_{C_i}}^\top}{\sqrt{d}}],\notag\\
&A = [\exp \frac{Qk_{1}^\top}{\sqrt{d}}, \ldots, \exp \frac{Qk_{l}^\top}{\sqrt{d}}], a_{C_{i}, j} = \exp\frac{Qk_{C_i, j}^\top}{\sqrt{d}},\text{ and } a_{j} = \exp\frac{Qk_{j}^\top}{\sqrt{d}}.\notag
\end{align}
We can restructure the computation hierarchically, first computing $V^h_{C_i}$ and $A^h_{C_i}$ for each context $C_i$:
\begin{align} 
V^h_{C_i} &= \mathrm{Softmax} \left( \frac{Q [k_{C_i, 1}^\top, \ldots, k_{C_i, l_{C_i}}^\top]}{\sqrt{d}} \right) \times [V_{C_i, 1}, \ldots, V_{C_i, l_{C_i}}], A^h_{C_i} = \mathrm{LogSumExp} \left( \frac{Q [k_{C_i, 1}^\top, \ldots, k_{C_i, l_{C_i}}^\top]}{\sqrt{d}} \right) 
\end{align}
Similarly, for the non-context tokens, we compute:
\begin{align} 
V^h &= \mathrm{Softmax} \left( \frac{Q [k_1^\top, \ldots, k_l^\top]}{\sqrt{d}} \right) \times [V_1, \ldots, V_l], A^h = \mathrm{LogSumExp} \left( \frac{Q [k_1^\top, \ldots, k_l^\top]}{\sqrt{d}} \right) 
\end{align}
After we get all these values, we can combine them while renormalizing with $A^h$:
\begin{align} 
O &= \mathrm{Softmax} \left( A^h_{C_1}, ..., A^h_{C_N}, A^h \right) \times [V^h_{C_1}, ..., V^h_{C_N}, V^h]
\end{align}
\subsection{Hierarchical Formula for APE.}

After incorporating all components in APE, we have a new $V^{h'}_{C_i}$ and $A^{h'}_{C_i}$ for each context $C_i$:
\begin{align} 
V^{h'}_{C_i} &= \mathrm{Softmax} \left( \frac{Q [k_{C_i, 1}^\top, \ldots, k_{C_i, l_{C_i}}^\top]}{T \cdot \sqrt{d}} \right) \times [V_{C_i, 1}, \ldots, V_{C_i, l_{C_i}}], A^{h'}_{C_i} = S \cdot \mathrm{LogSumExp} \left( \frac{Q [k_{C_i, 1}^\top, \ldots, k_{C_i, l_{C_i}}^\top]}{T \cdot \sqrt{d}} \right) 
\end{align}
For the non-context tokens, including our shared prefix, the formulas of $V^{h'}$ and $A^{h'}$ remain unchanged. Here, we introduce separate terms $V^{h'}_P$ and $A^{h'}_P$ for the shared prefix. Combining them, we have:
\begin{align} 
O &= \mathrm{Softmax} \left(A^{h'}_{P}, A^{h'}_{C_1}, ..., A^{h'}_{C_N}, A^{h'} \right) \times [V^{h'}_P, V^{h'}_{C_1}, ..., V^{h'}_{C_N}, V^{h'}]
\end{align}
\subsection{Relation with Equation~\ref{algo:ape}.}

Finally, we show that it can be rewritten as Equation~\ref{algo:ape},  with the only difference being that all contexts are treated as a single context. For an token from the position $j$ in context ${C_i}$, the final attention score $a''_{C_i, j}$ is
\begin{align} 
a''_{C_i, j} &= \frac{\exp(Qk_{C_i, j}^\top/T\sqrt{d})}{\sum_{n=1}^{N}\sum_{t=1}^{l_{C_n}}\exp(Qk_{C_i, t}^\top/T\sqrt{d})}\notag \\
&\cdot \frac{\exp\left(S \cdot \mathrm{LogSumExp} \left( \frac{Qt[k_{C_1, 1}^\top, \ldots, Qt[k_{C_1, l_{C_1}}^\top, \ldots, k_{C_n, 1}^\top, \ldots, k_{C_n, l_{C_n}}^\top]}{T \cdot \sqrt{d}} \right)\right)}{\exp\left(S \cdot \mathrm{LogSumExp} \left( \frac{Qt[k_{C_1, 1}^\top, \ldots, Qt[k_{C_1, l_{C_1}}^\top, \ldots, k_{C_n, 1}^\top, \ldots, k_{C_n, l_{C_n}}^\top]}{T \cdot \sqrt{d}} \right)\right) + \exp\left(\mathrm{LogSumExp} \left( \frac{Q [k_{1}^\top, \ldots, k_{l}^\top]}{T \cdot \sqrt{d}} \right)\right)} \\
&= \frac{\exp(Qk_{C_i, j}^\top/T\sqrt{d})}{\sum_{n=1}^{N}\sum_{t=1}^{l_{C_n}}\exp(Qk_{C_n, t}^\top/T\sqrt{d})} \cdot \frac{(\sum_{n=1}^{N}\sum_{t=1}^{l_{C_n}}\exp(Qk_{C_n, t}^\top/T\sqrt{d}))^S}{\sum_{n=1}^N (\sum_{t=1}^{l_{C_n}}\exp(Qk_{C_n, t}^\top/T\sqrt{d}))^S + \sum_{t=1}^{l}\exp(Qk_{t}^\top/\sqrt{d})} \\
&= \frac{\exp(Qk_{C_i, j}^\top/T\sqrt{d}) \cdot (\sum_{t=1}^{l_{C_i}}\exp(Qk_{C_i, t}^\top/T\sqrt{d}))^{(S-1)}}{(\sum_{n=1}^N \sum_{t=1}^{l_{C_n}}\exp(Qk_{C_n, t}^\top/T\sqrt{d}))^S + \sum_{t=1}^{l}\exp(Qk_{t}^\top/\sqrt{d})} = \frac{a'_{C_i, j}}{(\sum_{n=1}^{N}\sum_{t=1}^{l_{C_n}}a'_{C_i, t})^{S}+ \sum_{t=1}^{l}a_{t}}
\end{align}

This formula is equivalent to Equation~\ref{algo:ape}, except it combines the prefix and other non-context tokens for simplicity. Similarly, for the non-context tokens from position $j$, we can derive $a''_{j}$ as

\begin{align} 
a''_{j} = \frac{\exp(Qk_{j}^\top/\sqrt{d})}{\sum_{n=1}^N (\sum_{t=1}^{l_{C_n}}\exp(Qk_{C_n, t}^\top/T\sqrt{d}))^S + \sum_{t=1}^{l}\exp(Qk_{t}^\top/\sqrt{d})} = \frac{a_{j}}{(\sum_{n=1}^{N}\sum_{t=1}^{l_{C_n}}a'_{C_i, t})^{S}+ \sum_{t=1}^{l}a_{t}}
\end{align}

Combining these two components, we obtain the final formula presented in Equation~\ref{algo:ape}.

\subsection{Efficient Implementation.}

To combine the computation for context and non-context tokens, we employ flash attention twice—once for each part—and then merge the results. This only introduces a marginal computational overhead, as shown below.

\begin{lstlisting}[language=Python, numbers=none, basewidth={0.5em,0.5em}]
def ape_attention(query, key, value, temperature, scale):
    # split key and value states into context and non-context parts
    key_context, key_other = key
    value_context, value_other = value
    attn_output_context, lse_context = flash_attn(query, key, value, temperature = temperature)
    attn_output_other, lse_other = flash_attn(query, key, value)
    lse_context = lse_context*(scale)
    attn_weights = [lse_context, lse_other]
    attn_weights = Softmax(attn_weights)
    value_states = [attn_output_context, attn_output_other]
    attn_output = attn_weights @ value_states
\end{lstlisting}

\subsection{Future Directions.}

\begin{figure}[ht]
    \centering
    \includegraphics[width=\textwidth]{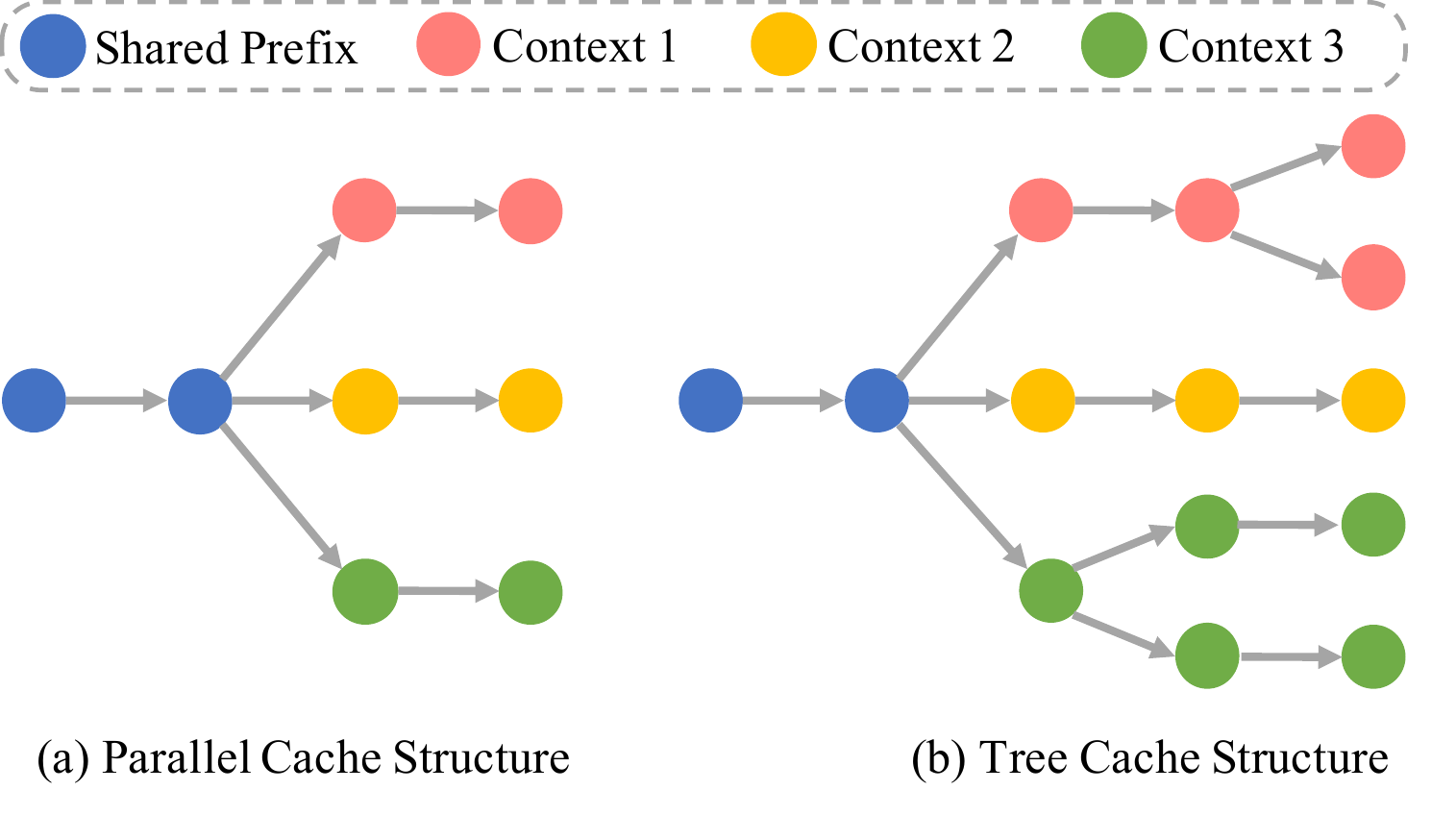}
    \caption{Beyond the parallel cache structure discussed in the main paper, APE can be extended to handle more complex cache structures from external sources, where each context forms a tree-like hierarchy. In this setup, computations can be performed hierarchically along each branch, progressively merging intermediate results into the final value state.} 
    \label{fig:app:structure}
\end{figure}

The hierarchical formulation of APE can naturally extend to more complex tree structures, as illustrated in Figure~\ref{fig:app:structure}. This flexibility allows each user query to be enriched with external knowledge organized in such structures, demonstrating APE's capability to handle structured external data effectively.  

\section{Comparing APE with Long-context LLMs.}
\label{sec:app:lclm}

\begin{table}[ht]
\centering
\caption{\textbf{Performance comparison between APE and long-context LLMs on LongBench~\citep{bai2023longbench}.}}
\resizebox{\linewidth}{!}{\setlength{\tabcolsep}{1mm}{
\begin{tabular}{l|cccccc}
\toprule
Method & NarratQA & Qasper & MultiFQA & GovReport & QMSum & LCC \\ \midrule
\textsc{LLaMA-3-8B-Instruct} & 19.32 & 32.83 & 43.38 & 27.89 & 22.40 & 53.22 \\
LLMLingua2 & 21.00 & 25.78 & 48.92 & 27.09 & 22.34 & 16.41 \\
StreamingLLM & 16.99 & 28.94 & 11.99 & 25.65 & 19.91 & 40.02 \\
Long-context FT & 14.88 & 21.70 & 47.79 & \textbf{32.65} & \textbf{24.76} & 55.12 \\
Self-Extend & 24.82 & 37.94 & 50.99 & 30.48 & 23.36 & 58.01 \\
\rowcolor{cyan!10}+APE & \textbf{26.87} & \textbf{39.14} & \textbf{59.12} & 29.10 & 23.08 & \textbf{66.09} \\
\midrule\midrule
Method & RepoBench-P & HotpotQA & 2WikiMQA & MuSiQue & MultiNews & Average \\ \midrule
\textsc{LLaMA-3-8B-Instruct} & 38.15 & 44.24 & 21.01 & 20.47 & 23.63 & 31.50 \\
LLMLingua2 & 20.56 & 40.16 & 24.72 & 20.85 & 21.34 & 26.29 \\
StreamingLLM & 26.16 & 32.76 & 20.12 & 17.32 & 21.49 & 23.76 \\
Long-context FT & 43.05 & 15.89 & 10.49 & 8.74 & 24.28 & 27.21 \\
Self-Extend & 41.83 & \textbf{51.09} & 24.17 & \textbf{28.73} & \textbf{24.11} & 35.96 \\
\rowcolor{cyan!10}+APE & \textbf{49.43} &50.11 & \textbf{28.06} & 25.79 & 22.40 & \textbf{38.11} \\
\bottomrule
\end{tabular}}}
\label{tab:app:single}
\end{table}

In Table~\ref{tab:app:single}, we further compare APE with Long-context LLM, including: (i) \textit{Prompt} \textit{Compression}: Truncation, LLMLingua2~\citep{pan2024llmlingua}, (ii) \textit{KV Cache Eviction}: StreamingLLM~\citep{xiao2023efficient}, (iii) \textit{Long-context FT}: Llama-3-8B-Instruct-262K~\citep{gradllama}, Llama-2-7B-Instruct-32K~\citep{togetherllama}, (iv) \textit{length extrapolation}: Self-Extend~\citep{jin2024llm}. Experimental results show that APE consistently outperforms all existing long-context LLM methods. We hypothesize that this improvement stems from APE enabling queries to access all past contexts, enhancing retrieval ability. However, since APE has limitations in identifying relationships between contexts, we do not emphasize its performance on current long-context tasks.

\section{APE Cache versus Prefix Cache}

\begin{figure}[ht]
    \centering
    \includegraphics[width=\textwidth]{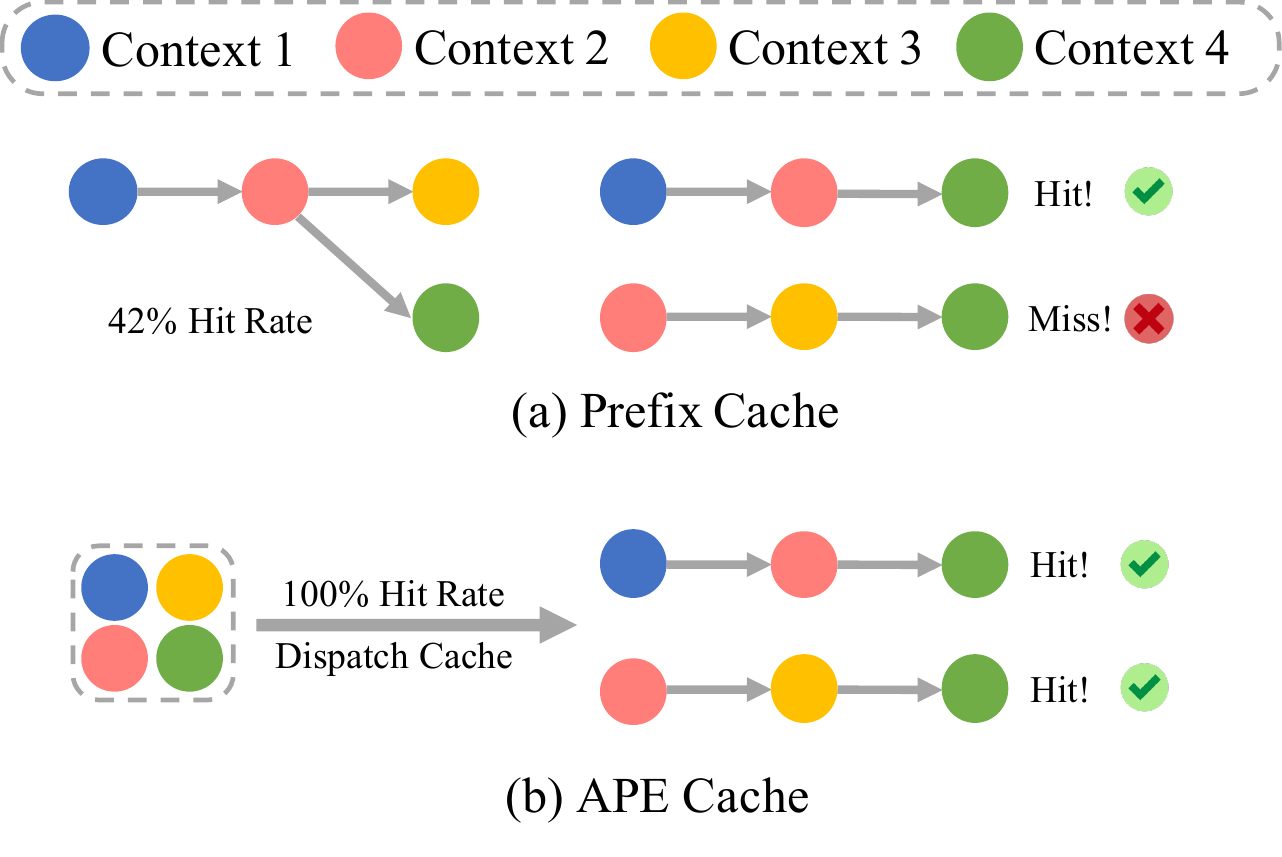}
    \caption{\textbf{Prefix Cache vs. APE Cache.} Our cache can keep a 100\% hit rate while the prefix cache only has 42\%.} 
    \label{fig:app:ape_cache}
\end{figure}

Finally, we compare the APE cache with the prefix cache to highlight our advantages in serving multiple queries within the CAG setting. Figure~\ref{fig:app:ape_cache} illustrates an example with four contexts where both caching strategies are allocated the same budget. Each query retrieves three contexts. Under these conditions, the prefix cache can only match a limited number of combinations, achieving an average hit rate of 41.7\%, whereas the APE cache ensures a 100\% hit rate. This gap will become even more pronounced as the number of contexts increases.

\end{document}